\documentclass[11pt]{article}
\usepackage[margin=1.3in]{geometry}  
\usepackage[parfill]{parskip}  
\usepackage{graphicx}
\usepackage{amsmath,amssymb,amsthm}
\usepackage[algoruled]{algorithm2e}
\usepackage{epstopdf}
\usepackage{xifthen}
\usepackage{enumitem}
\usepackage{todonotes}
\usepackage{mathtools}
\usepackage[displaymath, mathlines]{lineno}
\usepackage{bm}

\usepackage{sectsty}
\sectionfont{\fontsize{12}{15}\selectfont}
\subsectionfont{\fontsize{10}{15}\selectfont}
\subsubsectionfont{\fontsize{10}{15}\selectfont}
 \usepackage{booktabs}
  \usepackage{multirow}
  \usepackage{dsfont}
\usepackage[font=footnotesize]{caption}

\theoremstyle{plain}

  \newtheorem{theorem}{Theorem}[section]

\theoremstyle{definition}

\newcommand{\bbE}{\mathbb{E}}

\newcommand{\bbP}{\mathbb{P}}

\newcommand{\bbR}{\mathbb{R}}
\newcommand{\bbS}{\mathbb{S}}

\newcommand{\ba}{\boldsymbol{a}}
\newcommand{\bb}{\boldsymbol{b}}
\newcommand{\bc}{\boldsymbol{c}}

\newcommand{\bu}{\boldsymbol{u}}
\newcommand{\bv}{\boldsymbol{v}}
\newcommand{\bw}{\boldsymbol{w}}
\newcommand{\bx}{{\boldsymbol{x}}}

\newcommand{\bz}{\boldsymbol{z}}

\newcommand{\bA}{\boldsymbol{A}}
\newcommand{\bB}{\boldsymbol{B}}
\newcommand{\bC}{\boldsymbol{C}}

\newcommand{\bI}{\boldsymbol{I}}
\newcommand{\bJ}{\boldsymbol{J}}
\newcommand{\bK}{\boldsymbol{K}}
\newcommand{\bL}{\boldsymbol{L}}

\newcommand{\bQ}{\boldsymbol{Q}}

\newcommand{\bS}{\boldsymbol{S}}

\newcommand{\bU}{\boldsymbol{U}}

\newcommand{\bW}{\boldsymbol{W}}
\newcommand{\bX}{\boldsymbol{X}}

\newcommand{\bZ}{\boldsymbol{Z}}

\newcommand{\btheta}{\mathbb{\theta}}

\newcommand{\norm}[1]{\left\lVert#1\right\rVert}

\newcommand{\Tr}{\mathrm{Tr}}

\usepackage{comment}

\usepackage{subcaption}

\usepackage[style = alphabetic, maxnames=2, maxbibnames=99,uniquelist=false]{biblatex} 
\title{Theoretical Foundations of Representation Learning\\ using Unlabeled Data: Statistics and Optimization}

\author{Pascal Mattia Esser\footnote{Ludwig-Maximilians-Universit\"at M\"unchen, Department of Mathematics,\\
Akademiestraße 7, 80799 München, email: \href{mailto:pascal.esser@math.lmu.de}{pascal.esser@math.lmu.de}},
\and Maximilian Fleissner\footnote{Technical University of Munich, TUM School of Computation, Information and Technology,\\Boltzmannstr. 3, 85748 Garching,  email: \href{mailto:fleissner@cit.tum.de}{fleissner@cit.tum.de}},
\and Debarghya Ghoshdastidar\footnote{Technical University of Munich, TUM School of Computation, Information and Technology, 
\\ \quad Munich Data Science Institute (MDSI), 
Munich Center for Machine Learning (MCML),\\
Boltzmannstr. 3, 85748 Garching,   email: \href{mailto:ghoshdas@cit.tum.de}{ghoshdas@cit.tum.de}\\ 
\emph{Declaration: This preprint has been prepared while reporting the outcome of the research project ``Statistical Foundations of Unsupervised and Semi-supervised Deep Learning'' (GH-257/2-1) funded through the German Research Foundation's Prioirty Program on ``Theoretical Foundations of Deep Learning''. Hence, the preprint includes considerable focus on our own contributions during the project.}
}
}

\usepackage[colorlinks=true,allcolors=blue]{hyperref}%

\usepackage{nameref}

\begin{document}

\maketitle

\textbf{Abstract.} 
Representation learning from unlabeled data has been extensively studied in statistics, data science and signal processing with a rich literature on techniques for dimension reduction, compression, multi-dimensional scaling among others. However, current deep learning models use new principles for unsupervised representation learning that cannot be easily analyzed using classical theories. For example, visual foundation models have found tremendous success using self-supervision or denoising/masked autoencoders, which effectively learn representations from massive amounts of unlabeled data. However, it remains difficult to characterize the representations learned by these models and to explain why they perform well for diverse prediction tasks or show emergent behavior. To answer these questions, one needs to combine mathematical tools from statistics and optimization. This paper provides an overview of recent theoretical advances in representation learning from unlabeled data and mentions our contributions in this direction.

\section{Representation Learning: Past and Present}

Over the past decade, there has been a paradigm shift in learning representations from unlabeled data, where the focus has shifted from data compression to learning Euclidean representations of (potentially) unstructured data.   
Unsurprisingly, the evolution of the representation learning problem is aligned with the increasing focus on visual and text data, specifically the growing reliance on large language models and visual foundation models to tackle diverse applications in computational data science and beyond.
This paper discusses the theories of representation learning that are relevant in the current age of foundation models. 

Before discussing recent theoretical advances, it is helpful to reflect on the key difference between the classical and modern representation learning paradigms. One may identify two critical differences: (i) the prevalence of deep learning in current approaches towards representation learning and (ii) the context for learning representations of unlabeled data.
We elaborate further on both aspects with a focus on how they impact the development of a statistical theory for modern (unsupervised) representation learning.

\subsection{The importance of optimization in representation learning}
An introductory course in multivariate statistics often covers a range of textbook methods for unsupervised representation learning, including principal component analysis (PCA), independent component analysis, factor analysis, projection pursuit, non-negative matrix factorization, among others \cite{murphy_pml1Book,izenman2008modern,friedman1974projection}. 
Despite their prevalence in practice, a typical criticism of these approaches is that the learned representations are linear projections of the unlabeled data. 
There have been several attempts to generalize the principles of the aforementioned approaches to learn nonlinear representations. 
Notable classes of such methods are multidimensional scaling, including variants such as t-SNE \cite{tenenbaum2000global,izenman2008modern,van2008visualizing}, kernel methods \cite{scholkopf1998nonlinear}, tensor factorization \cite{kolda2009tensor}, and neural network-based approaches such as self-organizing maps, restricted Boltzmann machines (RBM), or autoencoders \cite{Goodfellow-et-al-2016,murphy_pml1Book}.

The popularity of deep neural networks in the early 2010s led to the dominance of deep learning models for unsupervised representation learning \cite{BengioCV13}. The shift in the research landscape is also influenced by the focus on image and text data, where it is known that unsupervised deep neural networks, such as RBMs, learn a hierarchy of visual features \cite{lee2011unsupervised}. 
Deep learning also provides practical convenience: 
\begin{quote}
    \textbf{Why is deep representation learning so prevalent?} 
    Unlike multivariate statistics principles, which often diverge in their practical implementations, current deep learning practices use a standardized framework to implement diverse ideas towards unsupervised representation learning. New approaches can be easily developed by applying standard optimization packages on a \emph{new loss function} and a \emph{new neural architecture}. 
\end{quote}
The unified implementation framework, compounded with the availability of more computational resources, has made deep learning the de facto model for representation learning, even beyond text or image domains---such as tabular data analysis \cite{ArikP21,YoonZJS20}, where matrix-based methods are historically preferred. 

Despite empirical success, deep representation learning has critical limitations. From a theoretical perspective, the key concern is the lack of rigorous understanding of the representations learned by the neural network-based models or the statistical properties of the trained models. 
An incomplete theory has severe practical implications. For example, design choices are often made based on empirical scaling laws \cite{HofmannEtAl22} that can be inaccurate, learned representations are inherently neither transparent nor interpretable \cite{abs-2108-07258,abs-2410-11444}, and the models are prone to adversarial attacks \cite{hendrycks2021unsolved}.

The challenges in developing a theoretical foundation for deep learning are not restricted to the context of representation learning. It is widely acknowledged that classical theories of statistical generalization cannot explain the performance of supervised deep neural networks \cite{ZhangBHRV21}. 
The principal bottleneck in providing a rigorous statistical understanding of deep learning lies in the complexity of the loss landscape of neural networks and the difficulty in characterizing the model learned via optimization.
A key focus of recent theoretical research has been the integration of the optimization and statistical aspects of deep learning to provide a better characterization of the generalization properties of trained neural networks \cite{Belkin21}. 

\subsubsection*{Open questions on optimization for deep representation learning.} 

Since the 1990s, there have been efforts to precisely characterize the dynamics of gradient descent in two-layer linear neural networks and to identify critical points of the dynamical system \cite{BALDI1989NN,Fukumizu98}. 
However, research in the past decade has provided a more precise description of the training dynamics of deep linear networks, typically in supervised settings, and the implicit biases of gradient-based optimization \cite{SaxeMG13,soudry2018implicit}. The optimization dynamics remains complex for neural networks with non-linearities, although it is possible to identify the dynamics for infinitely wide networks, which closely reflects the learning dynamics of kernel machines under infinitesimally small step sizes of gradient descent---this is referred to as the \emph{neural tangent kernel regime} or \emph{lazy training regime} \cite{ChizatB20,JacotHG18,AroraDHLW19,0001ZB20}. 
Recent works on tensor programs \cite{YangH21} and dynamic mean-field theory \cite{BordelonAP24,abs-2405-15712} provide a more accurate description of asymptotic dynamics under larger step sizes.
However, existing theoretical studies on the optimization of deep neural networks focus mainly on supervised learning settings, specifically involving squared, logistic, or hinge losses \cite{0001ZB20,ChizatB20,ChenHNW21}.
While few works consider general loss functions \cite{YangH21,YangL21a,soudry2018implicit}, the analysis is restricted to only a few optimization steps or assuming strong data assumptions, such as data separability. This is a major gap in the theory of optimization in the context of learning representations from unlabeled data.
\begin{quote}

\textbf{Open questions.}
What are the dynamics and implicit biases induced by optimization in the context of unsupervised representation learning? 
Specifically, how are the learning dynamics influenced by design choices common in representation learning, including architectures (bottleneck layers in autoencoders), type of loss functions (joint embedding loss used in self-supervised foundation models), nature of unlabeled data (masking or other augmentation), etc.?
\end{quote}
The present paper addresses some of these questions, but with a special focus on their impacts on statistical properties.

\subsubsection*{Open questions on statistical properties of trained neural networks}

The fundamental techniques of learning theory, such as universal approximation or uniform convergence bounds \cite{DeVoreHP21,Belkin21}, typically provide conservative guarantees for the generalization error of trained models and do not capture empirical trends \cite{ZhangBHRV21}. 
Precise generalization error curves for simple supervised models have been studied since the late 1990s \cite{SollichH02}, but work in the last decade has accurately identified the connections between training and generalization of learned models. For example, the bias of gradient descent towards small norm solutions causes the \emph{double descent phenomenon} in over-parameterized models \cite{Belkin21,MeiM22} or in the presence of early stopping \cite{HeckelY21}. Similarly, implicit regularization induced by dimension reduction or optimization (Landweber iterations) achieves minimax optimal generalization error rates \cite{DickerFH17}.

To this end, the neural tangent kernel regime provides the most generally applicable technique for analytically obtaining exact generalization error curves for trained neural networks \cite{MeiM22,BordelonCP20}, associated neural scaling laws \cite{BahriEKLS24}, and robustness of models \cite{BombariKM23,SabanayagamGGG25}. 
However, kernel approximations are inaccurate for feedforward neural networks with finite-width and transformer/attention-based architectures. In such cases, more precise results on the generalization error can be obtained by accounting for the training of the hidden layers or the attention layers, typically known as \emph{feature learning} and \emph{attention learning}, respectively. This analysis is currently restricted to highly specialized supervised learning problems such as regression, Gaussian mixture classification, and sparse problems \cite{KarpWLS21,Fu00M23,ShiWL23,BaESWWY22}.
Similar analysis of generalization in unsupervised deep learning is quite limited, and key questions remain unanswered.
\begin{quote}
    
    \textbf{Open questions.}
    What are the statistical properties of unsupervised deep learning models?
    Are there provable benefits of deeper networks and attention mechanism over kernel methods in the context of learning representations from unlabeled data?
\end{quote}
Preliminary steps towards answering these questions are discussed later in this paper in the context of reconstruction-based and joint embedding-based techniques for representation learning.
However, a characterization of the generalization error in unsupervised representation learning, or generally the statistical performance of such models, is not possible without considering the practical relevance of such models. Next, we briefly review the context for modern deep representation learning models.

\subsection{The influence of unsupervised representations on predictions}

Traditionally, learning representations of unlabeled data are either for the purpose of data compression, which requires dimension reduction or matrix factorization techniques, or for visualization and exploratory data analysis, encompassing approaches for data clustering, multidimensional scaling, etc.
Hence, the corresponding statistical theory often focuses on questions related to the error in estimating the latent signal, clusters, etc., or the loss of information \cite{Giraud21}. 
Importantly, prediction and generalization are not typical concerns in traditional unsupervised learning.

The objective of unsupervised representation learning has evolved in the current age of foundation models, where one relies on huge amounts of unlabeled data to learn ``useful representation'' that can be used in prediction tasks. In this context, representation learning is used to potentially reduce the dependence on expensive data labeling processes. For example, in large language models, where one uses next token prediction or masked modeling to learn useful embedding of unlabeled text data \cite{DevlinCLT19,abs-2303-08774}. Similarly, it is known that the representations learned in visual foundation models are naturally suitable for object detection or image classification, even though representation learning does not use image labels \cite{CaronTMJMBJ21,BardesPL22}. 

\begin{figure}[ht!]
    \centering
    \includegraphics[width=\linewidth]{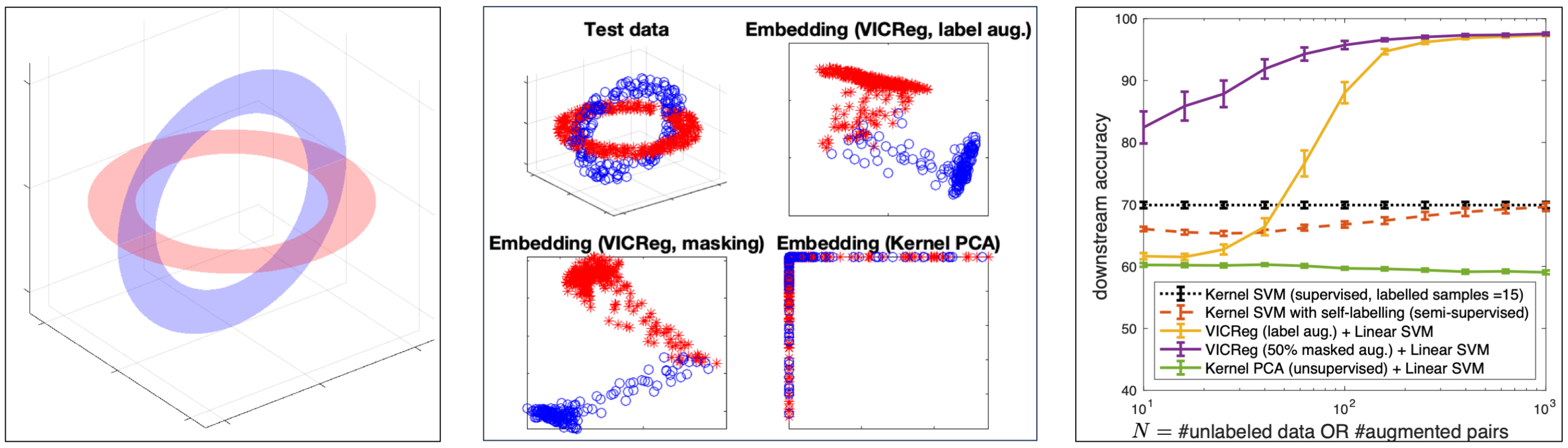}
    \caption{
    \textbf{Illustration of emergent properties of self-supervised representation learning in a simple setting.}
    The left plot shows the manifold of an example in $\mathbb{R}^3$, where we assume that data is generated from an union $2$ classes (two intersecting $2$-dimensional discs). Assume that one has access to 15 labeled samples and $N$ unlabeled samples with $N$ varying from $10$ to $1000$. The goal is to learn an ``informative representation'' from the unlabeled data so that a simple linear classifier, trained using only 15 labeled examples,  correctly predicts the class labels for new samples.
    \newline 
    We consider the case where one learns representations in $\mathbb{R}^2$ learned using the unlabeled $N$ samples.
    One could use traditional unsupervised methods like kernel principal component analysis (kernel PCA) \cite{scholkopf1998nonlinear}, or self-supervised techniques. We specifically consider a joint-embedding approach VICReg \cite{BardesPL22,CabannesKBLB23}, where for each unlabeled sample $x\in\mathbb{R}^3$, one generates a random augmented view  $x^+$ either by \emph{masking} each coordinate of $x$ with probability 0.5, or by \emph{rotating} the sample within the disc that it lies in (the latter is a hypothetical \emph{label-dependent augmentation} based on the philosophy that rotating images preserves their semantic meaning and the augmented data remains on same manifold).
    \newline
    The middle plot shows the embedding of 500 labeled test samples, where the representations $f:\mathbb{R}^3\to\mathbb{R}^2$ is learned with $N= 1000$ unlabeled examples. 
    The plot shows that $f (\cdot)$ from VICReg, with either augmentation, almost separates the two classes, whereas kernel PCA learns an uninformative representation.
    \newline
    The right plot shows the downstream predictive performance of unsupervised representation learning with varying $N$, averaged over $100$ independent runs. The downstream classifier is a linear support vector machine (SVM) trained on the representation $f (\cdot)$ of the 15 labeled samples. We also include two baselines that do not use  representation learning---a supervised kernel SVM trained on the 15 labeled samples in $\mathbb{R}^3$, and a semi-supervised approach of self-labeling, where kernel SVM predicts on available $N$ unlabeled data, and uses the pseudo-labels to update the model. 
    Supervised kernel SVM provides a baseline of 70\% accuracy, which does not improve with semi-supervised techniques.
    Unsupervised representation learning with kernel PCA does not learn ``more informative'' representations with more unlabeled data, whereas
    VICReg shows \emph{emergent behavior}---the downstream classification performance increases with the availability of large amount of unlabeled, augmented data. The improvement is particularly insightful for VICReg with label-dependent augmentation, whose performance is similar to kernel PCA when there are few unlabeled samples, but significantly improves when more unlabeled data is used.
}
    \label{fig:Emergent}
\end{figure}

Before proceeding to develop a statistical theory for modern representation learning, it is important to understand which aspects such a theory should address.
To this end, we note two interesting concepts that have been popularized in current practice.
\begin{itemize}
\item \textbf{Self-supervision} 

To learn useful representations from unlabeled data, it has become a common strategy to encode domain-specific knowledge through data augmentation. For example, the semantic meaning of an image does not change with rotations, and one learns representations that encode invariance to data augmentation \cite{BromleyGLSS93}. The practice of learning representations from unlabeled data in conjunction with domain-specific augmentations is broadly referred to as \emph{self-supervised learning}, which encompasses principles such as masked modeling \cite{DevlinCLT19}, denoising-based techniques \cite{VincentLLBM10}, joint embedding approaches \cite{BardesPL22}, contrastive learning \cite{ChenK0H20}, etc.

\item \textbf{Emergent property}

The use of large-scale data is critical to the success of foundation models. It has been empirically observed that the predictive performance of both language and visual foundation models drastically improves when a large amount of unlabeled data is used for self-supervised representation learning.
This phenomenon is often referred to as \emph{emergent property} of foundation models \cite{abs-2108-07258,CaronTMJMBJ21}. 
Figure~\ref{fig:Emergent} illustrates that the emergent property is an inherent characteristic of self-supervised representation learning and occurs even in simple examples. 
\end{itemize}
In view of the above discussion, one can pose the following concrete theoretical questions on representation learning in foundation models.
\begin{quote}
    
    \textbf{Open questions.}
    How can one derive guarantees on predictive performance of models that rely on self-supervised representation learning? What representations can be learned from the augmented data? What are the inductive biases of such models? Can one theoretically characterize emergent properties?
\end{quote}

\nocite{scholkopf1998nonlinear,CabannesKBLB23}

\clearpage

\subsection{Focus of this Work}

\begin{figure}[b!]
    \centering
    \includegraphics[width=\textwidth]{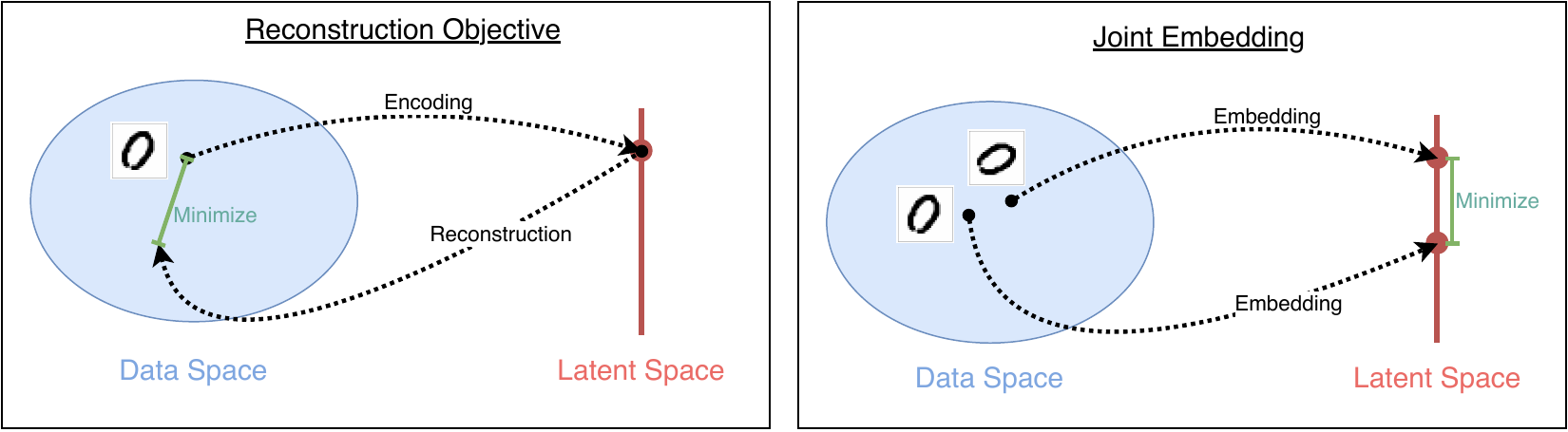}
    \caption{
    \textbf{Illustration of two principles for representation learning.}
    \emph{(Left) The objective of {reconstruction}}. An data instance is mapped into a lower dimensional latent space using an \emph{encoder function} and then mapped back to the original feature space using a \emph{reconstruction function}. The  functions are learned by minimizing the distance of the reconstruction from either the given instance or its augmentation.
    \emph{(Right) The objective of {joint embedding}.} This principle builds on the idea that semantically similar pair of data instances, usually obtained through data augmentation, should be embedded close to each other in the latent space. Hence, the embedding function is learned by minimizing the distance between the embedding augmented pairs of data instances, incorporating additional measures to ensure that trivial embeddings are not learned. 
    }
    \label{fig:data_setting}
\end{figure}

The broad aim of this paper is to discuss recent theoretical advances in characterizing the optimization and statistical properties of unsupervised and self-supervised representation learning, along with some discussion on their generalization and emergent properties.
Specifically, various results are presented on the dynamics of the learned representation under gradient descent, characterization of the learned representation, and the generalization error of the downstream predictors. 
The technical results are presented considering two specific principles for representation learning, illustrated in Figure \ref{fig:data_setting}:
\begin{itemize}
    \item \textbf{Reconstruction}

    In this original form, a reconstruction-based model learns a potentially low-dimensional representation through the process of reconstructing unlabeled examples. Autoencoder architectures are commonly used in these contexts, and if the activation of the hidden layer is linear, the learned representation is closely related to PCA \cite{BALDI1989NN}. 
    In the context of self-supervised settings, the data augmentation is in the form of introducing noise in unlabeled data or masking them, leading to \emph{denoising autoencoders} \cite{VincentLLBM10} and \emph{masked autoencoders} \cite{DevlinCLT19}, respectively.

    \item \textbf{Joint embedding}
    
    In its basic form, joint embedding aims to learn a joint representation of unlabeled augmented samples, such that the representation is invariant under augmentations. 
    Diverse formulations of this philosophy have been proposed, which can be broadly characterized into two principles. Non-contrastive learning aims to learn a representation that is identical for every augmented pair of samples \cite{BromleyGLSS93,BardesPL22,CaronTMJMBJ21}, while contrastive learning additionally uses non-augmented pairs to ensure that a nontrivial representation is learned \cite{ChenK0H20}. 
\end{itemize}
Sections 2 and 3, respectively, focus on reconstruction-based approaches based on autoencoders, and joint embedding approaches, including contrastive and non-contrastive methods. 
Section 4 discusses some results on the generalization error bounds for downstream prediction tasks using learned representations. 
Finally, we provide an outlook and key future research directions in Section 5.

In the subsequent discussion, we use the following notation.
We denote matrices by bold capital letters $\bA$, vectors as $\ba$, and $\bI_m$ for an identity matrix of size $m \in \mathbb{N}$. 
We assume that the $N$ data points $\{\bx_1\cdots\bx_N\} \in \bbR^d$ are collected in a data matrix $\bX\in\bbR^{d\times N}$. 
We use $\Vert\cdot\Vert$ to denote the Euclidean or $L_2$-norm for vectors, the Frobenius norm for matrices, and the Hilbert-Schmidt norm for operators.
$O(\cdot),\Omega(\cdot),\Theta(\cdot)$ refers to asymptotic notation.
Other notation is introduced in the respective sections.

\section{Reconstruction-based methods using Autoencoders}

Autoencoders (AE) \cite{Kramer1991AIChE_AE} are prototypical examples of neural networks used for reconstruction. Formally, an AE architecture is a composition of an encoder function $f:\bbR^d\to\bbR^k$ followed by a decoder function $g:\bbR^k\to\bbR^d$, where $\bbR^d$ is the original feature space and $\bbR^k$ is the latent space.
Traditionally, this setup has been used to learn to reconstruct data $g(f(\bx)) \approx \bx$. 
Given an unlabeled data matrix $\bX = [\bx_1, \bx_2, \ldots, \bx_N]$, the most common approach to learning the encoder and decoder is to minimize the reconstruction error. 
\begin{align}\label{eq: reconstruction}
   \min_{f,g} ~\sum_{i=1}^N \Vert \bx_i - g(f(\bx_i))\Vert^2.
\end{align}
Alternative formulations are also used in practice, including minimizing energy functions \cite{RanzatoPCL06} or imposing additional unsupervised tasks, such as clustering, on the learned representations \cite{yang2017towards}. 
If the dimension of the latent space or the \emph{bottleneck layer} $k<d$, then the obtained representation $f(\bx)\in\bbR^k$ provides a \emph{compression} of the data $\bx\in\bbR^d$. 
The representations learned by a trained encoder $f:\bbR^d\to\bbR^k$ have found success in a wide range of tasks, making AEs one of the popular deep learning architectures \cite{buades2005review,BengioCV13,yang2017towards}.


One of the fundamental theoretical questions is to characterize the representation $\bx \mapsto f(\bx)$ learned by optimization.
To this end, it is useful to consider a linear AE with a single hidden layer (bottleneck), that is, $f(\bx) = \bW_1\bx$ and $g(f(\bx)) = \bW_2f(\bx)$, where $\bW_1 \in \bbR^{k\times d}$ and $\bW_2 \in\bbR^{d\times k}$ are trainable weight matrices. 
Hence, one may replace Equation~\eqref{eq: reconstruction} by the following optimization:
\begin{align}
    \label{eq: linearAE}
     \min_{\bW_1, \bW_2} ~\norm{{\bX} - \bW_2 \bW_1{\bX}}^2 + \lambda \cdot r (\bW_1,\bW_2)
\end{align}
where $r$ denotes a regularization on the weight matrices, and $\lambda>0$ is a constant.
The training dynamics and the critical points for linear AEs can be characterized. 
In particular, \cite{BALDI1989NN} show that linear AE without regularization is equivalent to PCA---it finds solutions in the principal component that spans the subspace, but the individual components and the corresponding eigenvalues cannot be recovered.
\cite{Kunin20219pmlr_AE_losss_landscape} show that standard $l_2$ regularization reduces symmetry solutions to the group of orthogonal transformations.
Finally, \cite{Bao2020Neurips_AE_PC} show that nonuniform $l_2$ regularization allows linear AE to recover ordered, axis-aligned principal components.
Beyond linear AEs, \cite{RefinettiG22} provides the learning dynamics when the encoder $f$ is a nonlinear function.

\begin{figure}[b!]
\includegraphics[width=\linewidth]{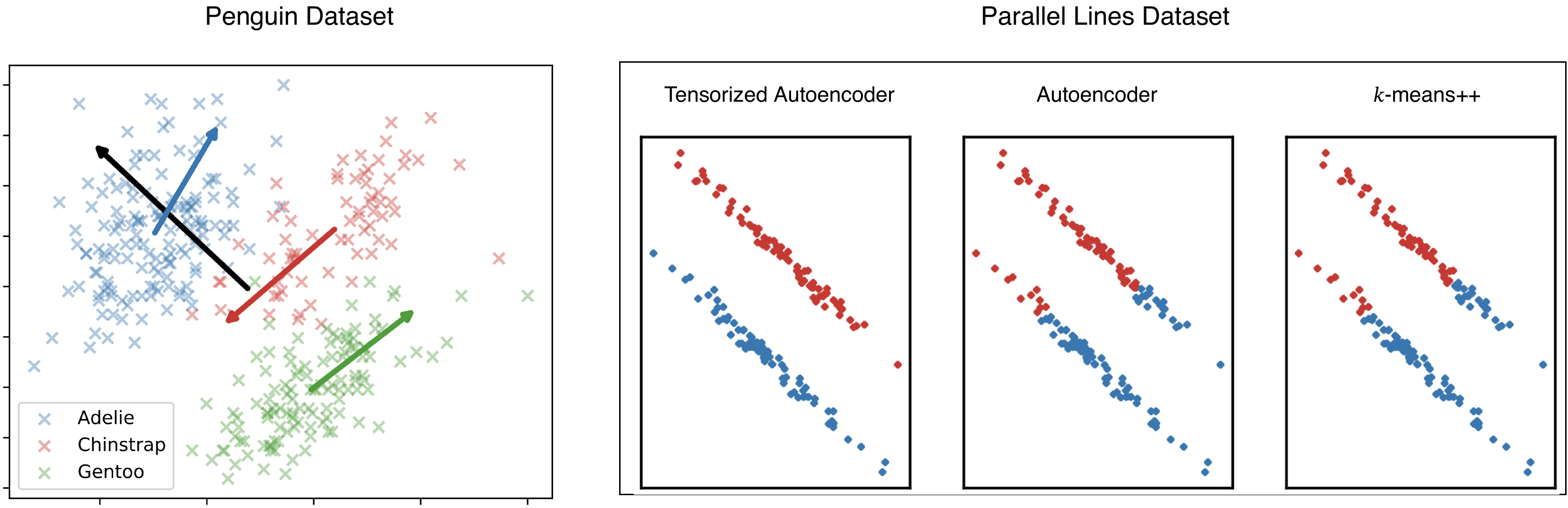}
  
    \caption{Source \cite{EsserMSG23}.
    \emph{(Left) Illustration of the Simpson's paradox \cite{Simpson1951stat}.} The scatter plot shows the dataset of 2-dimensional features for three different species of penguins \cite{Penguins}. The three clusters, for the different species, and their first principal component are plotted in red, blue and green, respectively.  
     A linear AE or PCA into $k=1$ dimensional latent space can only recover the principal component of the full dataset (shown in black), but cannot capture the characteristics of the individual species (clusters).
     Such examples of Simpson's paradox can only be found in non-linear models and real-world applications such as social or science and medical science. 
     \newline
     \emph{(Right) Performance of different clustering algorithms on Simpson's paradox data.}
     We consider a synthetic version of Simpson's paradox with noisy samples from two parallel lines in $\bbR^2$.
     One can either apply $k$-means++ on the original data (right), or on learned representations $f:\bbR^2\to\bbR$.
     AE recovers the principal component of the full data, which does align with the direction of the clusters. Hence, clustering the representations from AE does not recover the true clusters (middle).
     \cite{EsserMSG23} introduce tensorized AE, which learns representations for individual clusters, and results in better clustering performance. 
    }
    \label{fig:simpsonsParadox}
\end{figure}
\nocite{Simpson1951stat,Penguins,Clifford1982,Holt2016}

\begin{quote}

\textbf{Does data compression lead to good representations?}
Since the modern focus of the representation learning extends beyond data compression, it is natural to question whether reconstruction-based AEs (or nonlinear extensions of PCA) learn useful representations.
Figure~\ref{fig:simpsonsParadox} illustrates examples where standard AEs are not suitable and alternative principles are needed.
\end{quote}
In this paper, we will consider and theoretically analyze two alternatives to simple reconstruction losses of the form Equation~\eqref{eq: reconstruction}.
\begin{enumerate}
    \item \textbf{Reconstruction objectives based on augmented data.} 
    
    Current AE-based representation learning models go beyond mere input reconstruction and define self-supervised objectives using augmented data. 
    The most prominent self-supervised tasks involve denoising \cite{VincentLLBM10}, where $\bx^+$ is obtained by adding random noise to $\bx$, and masking \cite{DevlinCLT19,HeCXLDG22}, where parts of $\bx$ are removed or masked to construct $\bx^+$. 
    Subsequently, we present some theoretical results for denoising autoencoders (DAEs) \cite{buades2005review}, where the input data $\bX$ are corrupted by noise, which the model needs to ignore when reconstructing $\bX$. In Section~\ref{sec: AE dynamics}, we present an analysis of the learning dynamics and characterization of the generalization error of linear DAEs, proposed in \cite{abs-2505-24668}.

    \item \textbf{Generalized representations and new AE architecture.} 

    An alternative approach to learning better representations is to generalize the notion of a representation. Unlike standard AE that learns a single encoding $\bx \mapsto f(\bx)$, the idea is embed different inputs differently.
    In Section~\ref{sec: TAE}, we discuss a specific architecture of tensorized AE \cite{EsserMSG23}, where the bottleneck layer can learn multiple representations. Similar philosophy also lies behind deep clustering networks \cite{yang2017towards}.
\end{enumerate}

\subsection{Learning dynamics and generalization error of denoising AEs}\label{sec: AE dynamics}

Denoising autoencoders (DAEs) are trained to denoise unlabeled input data, but it is widely acknowledged that the resulting trained encoder learns useful low-dimensional representations \cite{vincent2010stacked}, which leads to its popularity in visual data analysis. 
For the purpose of theoretical analysis, we restrict the discussion to linear DAEs with $f(\bx) = W_1\bx$ and $g(f(\bx)) = W_2f(\bx)$, where the training objective is formulated as
\begin{align}\label{eq:DAE_linear}
    \min_{\bW_1, \bW_2}
    ~\mathcal{L}_{\text{trn}}(\bW_1, \bW_2) = \big\| {\bX} - \bW_2 \bW_1{(\bX + \bA)} \big\|^2 + \lambda \cdot \| \bW_2 \bW_1 \|^2,
\end{align}
where $\bA \in \bbR^{d\times N}$ denotes a noise matrix and 
the regularization of the form $r(\bW_1,\bW_2) = \Vert \bW_2\bW_1\Vert^2$. 
In \eqref{eq:DAE_linear}, one could denote $\bW_* = \bW_2\bW_1$ and optimize only over $\bW_*$. This results in a \emph{linear denoiser} \cite{SonthaliaN23,KausikSS24}, which can be solved as multivariate ridge regression. Although the statistical properties of ridge regression have been well studied, typical results assume additive noise is in the output \cite{Belkin21,DickerFH17}.
In contrast, a precise characterization of the generalization error of linear denoisers requires the additive noise to be taken into account. We refer to \cite{SonthaliaN23,KausikSS24} for the generalization error of linear denoisers, which exhibit interesting double-descent behavior in the over-parameterized regime.

In the subsequent discussion, we focus on the practically relevant setting where $\bW_1,\bW_2$ are separately trained and the bottleneck layer is of dimension $k < d$.
Studying this linear DAE with bottleneck allows us to precisely investigate the influence of the bottleneck dimension on the learned representations.
We consider the over-parameterized (or high-dimensional) regime, where the number of input observations is dominated by the dimensionality of the features, $c := \frac{d}{N} > 1$. 
For ease of exposition, it is convenient to assume that $\bX + \bA$ has full rank and the matrices $\bX\bX^\top$ and $\bA\bA^\top$ have simple eigenvalues.
Under this setting, one can rely on the classical result of \cite{BALDI1989NN} on critical points for linear neural networks, which leads to the following result.

\begin{theorem}[Global minimizer of linear DAEs \cite{abs-2505-24668}]\label{theorem:DAE_noskip_solution}
    Consider Equation~\eqref{eq:DAE_linear} in the ridgeless limit $(\lambda\to0)$, and denote $\bW_* = \bW_2 \bW_1$. The global minimizer $\bW_*$ of Equation~\eqref{eq:DAE_linear} converges to $\bW_* = P_{[k]}(\bX)(\bX + \bA)^{\dagger}$, where $P_{[k]}(\bX)$ is the rank-$k$ approximation to $\bX$ and $\dagger$ denotes the Moore-Penrose pseudoinverse.
\end{theorem}

It is important to note that the above solution is distinct from PCA, due to the noise $\bA$ inside the pseudoinverse, and also distinct from linear denoisers due to the rank-$k$ approximation. 
Using Theorem \ref{theorem:DAE_noskip_solution}, it is possible to characterize the generalization error of DAEs, under mild assumptions about the data-generating process. For training, we consider features
$\bX \in \mathbb{R}^{d \times N}$ and an additive Gaussian noise matrix $\bA \in \mathbb{R}^{d \times N}$ with entries sampled independently from $\mathcal{N}\left(0, \frac{\eta_{\text{trn}}^2}{d}\right)$. 
We study generalization in terms of the expected denoising error on test data
\begin{align}
    \mathcal{L}_{\text{tst}}(\bW_*) := \frac{1}{N_{\text{tst}}} \mathbb{E}_{\bA_{\text{trn}}, \bA_{\text{tst}}} \left[\|\bX_{\text{tst}} - \bW_{*} (\bX_{\text{tst}} + \bA_{\text{tst}})\|^2\right],
\end{align}
$\bX_{\text{tst}} \in \mathbb{R}^{d \times N_{\text{tst}}}$ is the test data, perturbed by Gaussian noise matrix $\bA_{\text{tst}} \in \mathbb{R}^{d \times N_{\text{tst}}}$, that is,  the entries of $\bA_{\text{test}}$ are $\mathcal{N}\left(0, \frac{\eta_{\text{tst}}^2}{d}\right)$ and $\eta_{\text{trn}}, \eta_{\text{tst}} = \Theta(1)$. 

For the following results, assume that the training data $\bX$ and the test data $\bX_{\text{tst}}$ lie in the same low-dimensional subspace of dimension $r \ll d,N$. Formally, the test data $\bX_{\text{tst}}$ satisfy $\bX_{\text{tst}} = \bU \bL$, for $\bU \in \mathbb{R}^{d \times r}$ the left singular vectors of $\bX$ and for some non-zero coefficient matrix $\bL \in \mathbb{R}^{r \times N_{\text{tst}}}$. The low-rank assumption on the data is well supported by empirical evidence that real-world data sets are approximately low-rank, as argued in \cite{udell2018bigdatamatricesapproximately}.
In addition, we assume that $\|\bX\|_2 = \Theta(1)$ and the ratio between the largest and the smallest nonzero singular values of $\bX$ is $\Theta(1)$. With these assumptions in place, we may state the following.

\begin{theorem}[Generalization error of over-parametrized linear DAEs \cite{abs-2505-24668}]\label{theorem:test_risk_noskip}
    Let $\sigma_i$ denote the $i$-th singular value of $\bX$, and define $\alpha_i := \sigma_i \eta_{\text{trn}}^{-1}$.
    Let $d \geq N + r$, and $c := \frac{d}{N}$. Let $\bJ \in \mathbb{R}^{r \times r}$ be the diagonal matrix with
    $\displaystyle\bJ_{ii} = \big(\alpha_i^2 + 1\big)^{-2} \cdot \mathds{1}_{i \in [k]} 
             + \mathds{1}_{i \notin [k]}$,
    where $\mathds{1}_{(\cdot)}$ denotes the indicator function.
    Then,
    \begin{align*}
    \mathcal{L}_{\text{tst}}(\bW_*) = \frac{1}{N_{\text{tst}}} \Tr(\bJ \bL \bL^\top) + \frac{\eta_{\text{tst}}^2 c}{d(c - 1)} \sum_{j \in [k]} \frac{\alpha_j^2}{1 + \alpha_j^2} + O\left(\frac{1}{d^2}\right).
    \end{align*}
\end{theorem}

\begin{figure}[ht]
    \centering
    {{\includegraphics[height=3.0cm]{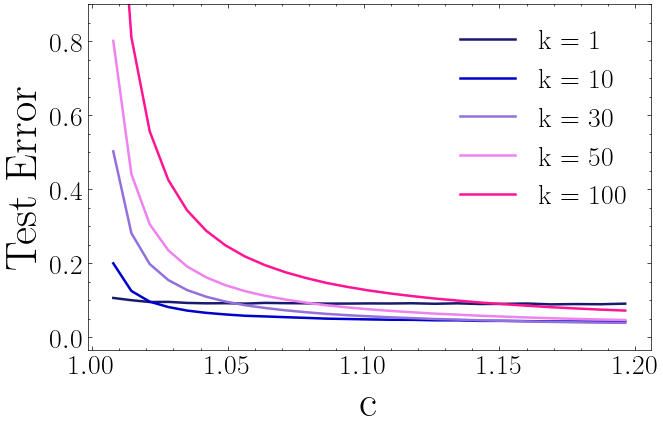} }}%
    {{\includegraphics[height=3.0cm]{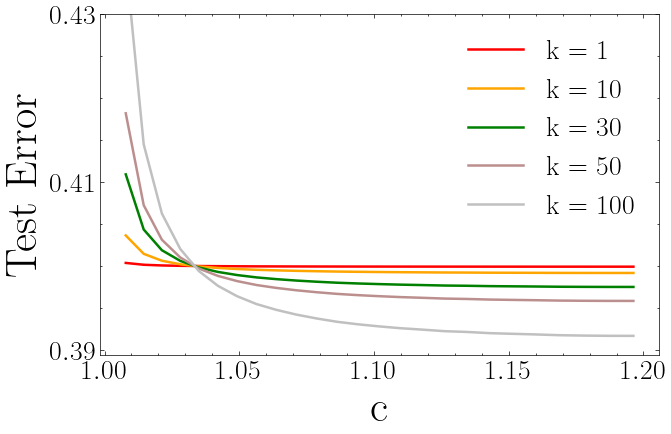} }}%
    {{\includegraphics[height=3.0cm]{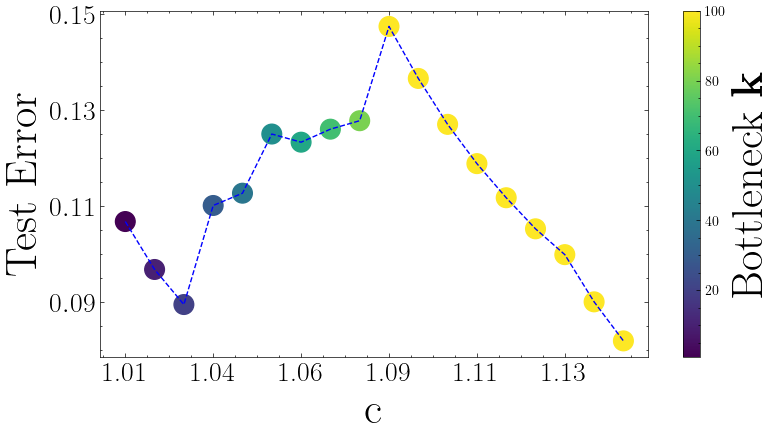} }}%
\caption{Source \cite{abs-2505-24668}. \textbf{Impact of bottleneck dimension of linear DAE on generalization error.} We plot the test errors of linear DAEs with and without skip connection on CIFAR-10, illustrating how the bottleneck dimension $k$ and $c=\frac{d}{N}$ jointly influence generalization. For the experiments, each sample was reshaped into a $3072$-dimensional vector, and the rank of the dataset was set to $r=100$ using SVD. Since the dataset has a fixed ambient dimension $d$, our numerical experiments focus on varying the number of training samples $N$. 
\emph{(Left \& Middle)} The left and middle plot show the denoising error on test data with varying $c = \frac{d}{N}$ for the linear DAEs without and with skip connections, respectively. Both plots demonstrate that the optimal choice of $k$ depends on the over-parameterization ratio $c$, reflecting a distinct bias-variance trade-off in different regimes.
\emph{(Right)} To study the impact of over-parameterization, the right plot is constructed by jointly increasing both $k$ and $c$ in the model without skip connections. In particular, this plot demonstrates that jointly increasing $c$ and bottleneck dimension $k$ leads to a \textit{second peak} in the test curve within the overparameterized regime.
}
\label{fig:bottleneck}
\end{figure}

Since $\bL$ depends only on the test data, the overall magnitude of $\Tr(\bJ \bL \bL^\top)$ is mainly influenced by the size of the diagonal entries of $\bJ$. Thus, the first term \textit{decreases} as bottleneck dimension $k$ increases toward $r$. In contrast, the second term \textit{increases} as $k$ grows. This trade-off behavior aligns exactly with the \textit{classical understanding of the bias--variance trade-off}. In fact, one can show that the first term is exactly the bias of $\bW_*$ whereas the second is the variance of the predictor. 
In particular, the aforementioned trade-off emerges in the over-parameterized regime, where $d>N$. 
Figure \ref{fig:bottleneck} (left) shows the generalization error for linear DAE as a function of the ratio $c = \frac{d}{N}$. We observe the typical phenomenon that over-parameterization leads to a decrease in generalization error as well-known in the \emph{double descent literature} \cite{Belkin21}.
However, in the case of the DAE, the complexity of the model is controlled by the width of the bottleneck $k$, not only the input dimension.

It is further useful to study a practical variant of DAE, where a skip connection is added across the bottleneck layer resulting in the following optimization. 
\begin{align}\label{eq:DAE_linear_skip}
    \mathcal{L}_{\text{trn}}(\bW_1, \bW_2) = \| {\bX} - (\bI + \bW_2 \bW_1){(\bX + \bA)} \|^2 + \lambda \cdot \| \bW_2 \bW_1 \|^2.
\end{align}
Here $\bI$ is the $d$-dimensional identity matrix that maps inputs directly to the output layer, without compressing them in the bottleneck. 
Although adding a skip connection is counterintuitive for AEs,
practical implementations such as the U-Net architecture \cite{ronneberger2015u} routinely incorporate skip connections as a core architectural feature. 
It is an open question to understand the impact of skip connection on the generalization error in DAEs. We extend Theorems \ref{theorem:DAE_noskip_solution}--\ref{theorem:test_risk_noskip} to address this question.

\begin{theorem}[Global minimizer and generalization error of linear DAE with Skip Connection \cite{abs-2505-24668}]\label{theorem:test_risk_skip}
    In the ridgeless limit $\lambda \rightarrow 0$, the global minimizer $\bW_*$ of Equation~\eqref{eq:DAE_linear_skip} approaches $\bW_* = - P_{[k]}(\bA)(\bX + \bA)^{\dagger}$.
    Furthermore, under similar assumptions as Theorem \ref{theorem:test_risk_noskip}, the generalization error $\mathcal{L}_{\text{tst}}$ is given by
    \begin{align*}
        \mathcal{L}_{\text{tst}}(\bW_*)
        = \eta_{\text{tst}}^2 \left(1 - \frac{k}{d}\right) &+ \frac{k}{d N_{\text{tst}}} \Tr\left(\bJ^{\text{sc}}\bL \bL^\top\right) + \frac{\eta_{\text{tst}}^2 k}{d^2} \frac{c}{c - 1}\sum\limits_{i=1}^r \frac{\sigma_i^2}{(\eta_{\text{trn}}^2 + \sigma_i^2)} \nonumber \\
        &+ \frac{3\eta_{\text{tst}}^2 k}{dN} \frac{1}{c} \sum\limits_{i=1}^r \frac{\eta_{\text{trn}}^2 \sigma_i^2}{(\eta_{\text{trn}}^2 + \sigma_i^2)} + O\left(\frac{1}{dN_{\text{tst}}}\right).
    \end{align*}
    where $\bJ^{\text{sc}}$ is a diagonal matrix with $\bJ^{\text{sc}}_{ii} = \frac{c + (c - 1)\sigma_i^2}{c(1 + \eta_{\text{trn}}^{-2} \sigma_i^2)^2}$ for each $i \in [r]$. 
\end{theorem}

It is useful to compare Theorem \ref{theorem:test_risk_noskip} with Theorem \ref{theorem:test_risk_skip} (also see Figure \ref{fig:bottleneck}). Firstly, observe that the variance term in the former is responsible for the sharp increase in the test curves as the ratio 
$c$ approaches $1$, due to the term $(c - 1)^{-1}$. A similar, but less pronounced trend is observed in the model with skip connections, where the term
$\frac{\eta_{\text{tst}}^2 k}{d^2} \frac{c}{c - 1}\sum_{i=1}^r \frac{\sigma_i^2}{(\eta_{\text{trn}}^2 + \sigma_i^2)}$ also includes the factor $(c - 1)^{-1}$. However, in contrast to the model without skip connections, the expression is multiplied with an additional factor of $d^{-1}$. This suggests that \textit{skip connections help mitigate the sharp rise in variance} that typically occurs when the model is in the moderately over-parameterized regime, leading to more stable generalization performance.

\subsection{Learning multiple representations with tensorized autoencoders}\label{sec: TAE}

We now consider an example where the AE architecture is generalized to incorporate additional structures in the data. Conceptually, the broad aim is to learn representations that preserve important structures of the data, while removing noise dimensions. 
Linear AE or PCA retains only the high-variance directions, but it could sometimes be more relevant to preserve other topological properties. 
In this section, we restrict ourselves to the case where the goal is to learn latent cluster representations, which can also be used to cluster the data (see Figure \ref{fig:simpsonsParadox}).

In \cite{EsserMSG23}, we introduce a modified AE architecture that we term \emph{Tensorized Autoencoders (TAE)}. At a high level, the TAE architecture consists of $m$ AEs in parallel. Given a data set $\{\bx_i\}_{i=1}^N$, each input $\bx_i$ is passed to $j^{th}$ AE with a nonnegative weight $\bS_{j,i}$. This allows us to learn $m$ different latent representations, one from each AE.

Assuming that a data set has $m$ linearly separable clusters, one can show that tensorized linear AEs provably recover the principal directions of \emph{each cluster} while jointly learning the assignment of clusters.
To see this, we define two-layer linear TAEs formally as follows.
For each data point $\bx_i$, define $\widetilde{\bx}_i^{(j)}:=\bx_i - \bc_j$ as a centered data, where $\bc_1,\ldots,\bc_m$ are cluster centers learned defined later.
The $j$-th AE in the TAE is parameterized by the linear encoder map, $\widetilde{\bx}_i^{(j)} \mapsto \bW^{(j)}_1\widetilde{\bx}_i^{(j)}$, where $\bW^{(j)}_1\in\bbR^{k \times d}$ (we view the encoder as the embedding of the $j$-th cluster).
Similarly, the linear decoder map for $j$-th AE is parameterized by $\bW^{(j)}_2\in\bbR^{d \times k}$.
The TAE parameters are learned by solving the following optimization.
 \begin{align}\label{eq: relaxed cost}
     \min_{\{ \bW^{(j)}_2,\bW^{(j)}_1 \}_{j=1}^m, \bS} ~ &\sum^N_{i=1} \sum^m_{j = 1}\bS_{j,i}\Big[\norm{\widetilde{\bx}_i^{(j)} - \bW^{(j)}_2\bW^{(j)}_1\widetilde{\bx}_i^{(j)}}^2 ~ + ~ \lambda \norm{\bW^{(j)}_1\widetilde{\bx}_i^{(j)}}^2\Big], \\ \nonumber
     \text{s.t.~ }& \sum_{j=1}^m \bS_{j,i} = 1, 
     \quad \bS_{j,i} \geq 0,
     \quad \bW^{(j)}_1\bW^{(j)T}_1 = \bI_k,
     \quad \bc_j = \frac{\sum\limits_{i=1}^N \bS_{j,i} \bx_i}{\sum\limits_{i=1}^N \bS_{j,i}},
 \end{align}
where $\lambda>0$ is a regularization constant.
Interpreting $\bS_{j,i}$ as the probability that $\bx_i$ is from the $j$-th cluster, the above objective can be interpreted as minimizing the expected reconstruction error of the samples, regularized with a $k$-means clustering cost. 
The constraints on $\{\bS_{j,i}\}_{j=1}^m$ ensure that the weights are indeed probabilities and the orthogonality of $\bW_1^{(j)}$ ensures that the encoders are projections.
The following theorem characterizes the optimal parameters of a linear TAE, showing that linear TAEs indeed learn clustering-specific representations of the data.

\begin{theorem}[Parameterization at optimal for TAE \cite{EsserMSG23}]\label{Th: optimal relaxed}
For $0 < \lambda \leq 1$, optimizing Equation~\ref{eq: relaxed cost} results in the parameters at the optimum satisfying the following:
\begin{enumerate}
    \item
    {Class Assignment:} 
    Any optimal $\bS$ satisfies $\bS_{j,i} \in \{0,1\}$.
    In combination with the linear constraint, the above implies that each $\bx_i$ is assigned exactly to one cluster and is encoded by exactly one AE. 

    \item
    {Encoding / Decoding (learned weights):}  At optimality, $\bW^{(j)T}_2 = \bW^{(j)}_1$ correspond to the top $k$ eigenvectors of 
    $\hat{\mathbf{\Sigma}}_j: =\displaystyle \sum^N_{i=1}\bS_{j,i}\left(\bx_i - \bc_j\right)\left(\bx_i - \bc_j\right)^\top$. 
\end{enumerate}
As a consequence, for any optimal solution, $\bc_j$ and $\hat{\mathbf{\Sigma}}_j$ act as estimates for the means and covariances for each specific class, respectively.
\end{theorem}

The above theorem demonstrates the suitability of the TAE formulation. In practice, it is more useful to consider nonlinear AEs, $f_j(\cdot), g_j(\cdot)$, resulting in an optimization of the form:
\begin{align*}
    \min_{\{f_j, g_j\}^m_{j=1},\bS}\sum^N_{i=1} & \sum^m_{j = 1}\bS_{j,i}\left[\norm{\widetilde{\bx}_i^{(j)} - g_j\left(f_j\left(\widetilde{\bx}_i^{+(j)}\right)\right)}  + \lambda\cdot r(f_j,g_j)\right],
\end{align*}
where it is implicitly assumed that optimization is performed over a certain parametric form of the encoder $f_J$ and decoder $g_j$.
General regularization functions can be imposed on the AEs, but regularization based on $k$-means is popular in the deep clustering literature \cite{yang2017towards}.
Further empirical work also extends the idea of tensorization beyond AEs to tensorized variational AEs and restricted Boltzmann machines \cite{sabanayagam2024clusterspecificrepresentationlearning}.
However, it is challenging to characterize the optimal solution or the training dynamics beyond linear TAEs.

\section{Self-supervised joint embedding methods}

AE architecture and its variants focus on reconstructing or recovering the original data $\bx$. As a consequence, the learned representation $\bx\mapsto f(\bx)$ ignores the data features that have less impact on reconstruction. 
This may not be desirable if the representations are used for downstream prediction tasks.
One can see this in the following simple example.

\begin{quote}

\textbf{Example (AE may ignore direction relevant for Gaussian mixture classification)}
    Consider data sampled from Gaussian mixture distribution in $\bbR^2$
    \begin{align*}
        \bx = \left(\begin{array}{c}x^{(1)}\\x^{(2)}\end{array}\right) \sim 
    \frac12 \mathcal{N}\left( 
    \left[\begin{array}{c}-1\\0\end{array}\right], 
    \left[\begin{array}{cc}1&~0\\0&~3\end{array}\right]\right) + 
    \frac12 \mathcal{N}\left( 
    \left[\begin{array}{c}1\\0\end{array}\right], 
    \left[\begin{array}{cc}1&~0\\0&~3\end{array}\right]\right).
    \end{align*}
    If a linear AE is used with a bottleneck layer of width 1, then asymptotically as $n\to 0$, the AE learns the principal component $\bx \mapsto f(\bx) = x^{(2)}$. 
    However, if the objective is to use the representations for downstream classification, it is desirable to project the data on the axis that separates the classes, $\bx \mapsto x^{(1)}$.
\end{quote}
The above situation, where the useful features do not align with the features learned by reconstruction, is often encountered in visual foundation models \cite{balestriero2024learning}. In these cases, \textit{joint embedding methods} are preferred.
These methods rely on pairs $\bx,\bx^+ \in \bbR^d$ (or tuples) of semantically similar samples and learn a representation $f: \bbR^d \rightarrow \bbR^k$ such that $f(\bx), f(\bx^+)$ are ``close'' or ``aligned''.
The idea was introduced in \cite{BromleyGLSS93}, where a Siamese network--two neural networks with identical weights--was used on pairs of signatures from the same person for the application of signature verification.
In practice, semantically similar samples are generated through \emph{data augmentation}, which depends on the data domain. 
For example, image augmentation through random cropping, rotation, color jitter, etc. help to learn useful representations for image classification \cite{ChenK0H20}.
In current foundation models, labeled data are not used to learn the representation (often called \emph{self-supervised pretraining}). The hope is that the augmented pair $\bx,\bx^+$ belong to the same class for downstream prediction tasks. 
\begin{quote}
    
    \textbf{Examples (Data augmentation resulting in same or different class labels)}
     In the context of classifying cat images from dog images, augmentation such as cropping or rotation generate image with similar class labels \cite{ChenK0H20}.
     However, in the context of tumor detection in medical images, cropping could change the semantic meaning if the tumor is cropped \cite{HuangPJLYC23}.
\end{quote}
The astute reader will have noticed that it is trivially possible to perfectly align all pairs $f(\bx), f(\bx^+)$ in the representation space by learning a trivial constant function $f(\bx)$ (known as a \emph{collapse}).
Thus, a core component of joint embedding methods is to define a loss function $\mathcal{L}$ that prevents $f$ from learning such degenerate solutions. 

Given augmented pairs $\{(\bx_i, \bx_i^+)\}_{i=1}^N$, the \emph{Barlow Twins loss} function \cite{ZbontarJMLD21} pushes the cross-correlation matrix $\bC \in \bbR^{k \times k}$ between the pairs $f(\bx_i)$ and $f(\bx_i^+)$ towards the identity matrix $\bI_k$. Formally, for a hyperparameter $\lambda > 0$, we learn $f$ by minimizing
\begin{align}
\label{eq:BT}
    \mathcal{L}_{BT}(f) =
    \sum_{j=1}^k \left( 1 - \bC_{jj} \right)^2
    + \lambda \sum_{j \neq l} \bC_{jl}^2
\end{align}
where $\bC$ is the cross-correlation matrix between $\{ f(\bx_i) \}_{i=1}^N$ and $\{ f(\bx_i^+) \}_{i=1}^N$, that is
\begin{align*}
    \bC_{jl} = \frac{\frac{1}{N}\sum\limits_{i=1}^N f_j(\bx_i) f_l(\bx_i^+)}{\sqrt{\frac{1}{N}\sum\limits_{i=1}^N f_j(\bx_i)^2} \sqrt{\frac{1}{N}\sum\limits_{i=1}^N f_l(\bx_i^+)^2}}
\end{align*}
Intuitively, enforcing $\bC_{jj} \approx 1$ ensures that $f_j(\bx_i) \approx f_j(\bx_i^+)$ and thus preserves the invariances encoded in the pairs $\bx_i,\bx_i^+$. At the same time, decorrelating $f_j(\bx_i)$ and $f_j(\bx_i^+)$ ensures that each dimension learns a different set of features, preventing dimension collapse. Since cross-correlation complicates the mathematical treatment of the Barlow Twins loss, theoretical studies use the symmetrized cross-moment matrix \cite{SimonKLGFA23,fleissner2025infinite} 
\begin{equation}
 \bC= \displaystyle\frac{1}{2N} \sum_{i=1}^N \left( f(\bx_i)f(\bx_i^+)^\top + f(\bx_i^+)f(\bx_i)^\top\right)   
 \label{eq:symC}
\end{equation} 
to define $\mathcal{L}_{BT}$ in \eqref{eq:BT}.
An improvement over Barlow twins was proposed in \cite{BardesPL22}, called the \emph{Variance-Invariance-Covariance Reguralization} or simply the \emph{VICReg loss} function, which appends $\mathcal{L}_{BT}$ with an invariance term $\sum_i \Vert f(\bx_i) - f(\bx_i^+)\Vert^2$ to ensure that the embeddings of the augmented pairs are aligned. VICReg has been studied in few theoretical works \cite{CabannesKBLB23,feigin2024theoretical}.

Barlow Twins and VICReg models are called examples of \emph{non-contrastive learning} to distinguish them from \emph{contrastive learning}---another class of joint embedding methods, where trivial solutions are avoided by imposing that augmentations from different samples are embedded far apart in the representation space.
Two popular contrastive losses are SimCLR \cite{ChenK0H20} and spectral contrastive loss \cite{HaoChen2021ProvableGF}, where the latter is defined as follows.
\begin{align}
\label{eq:SCL}
\mathcal{L}_{SCL}(f)
&= -\frac2N\sum_{i=1}^n f(\bx_i)^\top f(\bx_i^+) + \frac{1}{N^2}\sum_{\substack{i,j=1 \\ i\neq j}}^N \left(f(\bx_i)^\top f(\bx_j^+)\right)^2 
\end{align}
The above loss can diverge to $-\infty$, which is prevented by projecting the representations onto the unit ball \cite{HaoChen2021ProvableGF}, or by regularizing the norm of the representation \cite{EsserFG24,feigin2024theoretical}.
There exist numerous empirical studies and heuristic explanations on what these loss functions learn, but they fall short of providing a principled scientific treatment of joint embedding methods, and several open questions remain. In this paper, we address the following questions.

\begin{itemize}
    \item[1.] \textbf{Characterizing optimal solution}  
    
    What are the representations or patterns learned by minimizing \eqref{eq:BT} or \eqref{eq:SCL}?
    
    \item[2.] \textbf{Generalization} 
    
    Do the learned patterns generalize to new data? Can we trust a learned $f$ to achieve a small $\mathcal{L}_{\text{BT}}$ or $\mathcal{L}_{\text{SCL}}$ loss on new samples?
    
    \item[3.] \textbf{Expressivity} 
    
    Which representations $f$ can be learned by minimizing $\mathcal{L}_{\text{BT}}$ or $\mathcal{L}_{\text{SCL}}$? What is the \emph{ideal} data augmentation that learns $f$?
    
    \item[4.] \textbf{Implicit bias of optimization} 
    
    If we solve the optimization problems in \eqref{eq:BT} or \eqref{eq:SCL} by gradient descent, then what are the resulting representations?
\end{itemize}
To answer the first three questions, it is convenient to assume that $f$ is learned using a kernel model instead of a neural network. Recall that for any positive definite kernel $\kappa : \bbR^d \times \bbR^d \to \bbR$, there exists a corresponding reproducing kernel Hilbert space (rkhs) $\mathcal{H}$ and a feature map $\phi: \bbR^d \rightarrow \mathcal{H}$ such that $\kappa(\bx,\bx')= \langle \phi(\bx), \phi(\bx')\rangle$  \cite{ScholkopfMIT}.
A kernel model for representation is a function of the form $f(\bx) = \bW \phi(\bx)$, where the linear operator $\bW: \mathcal{H}\to\bbR^k$ is learned to optimize an objective. 
When $\mathcal{H}$ is finite-dimensional, say $\mathcal{H} = \bbR^p$, then $\bW \in \bbR^{k \times p}$ is simply a matrix with $k$ rows, each defining one of the $k$ output dimensions of the function. The kernel setting comes with several advantages. 
Closed-form expressions of the optimal solution $f$ can be derived for kernel models, which simplifies questions on expressivity and deriving ideal augmentation. The kernel literature also provides several techniques to estimate the generalization error of the learned $f$. Finally, we discuss that 
if a wide neural network is used to learn the representations,  then the implicit bias of gradient descent results in a solution at convergence that is close to a learned kernel model.

\subsection{Neural tangent kernel regime for joint embedding models}

As discussed in Section 1, the \emph{neural tangent kernel (NTK) regime} is well known in the supervised deep learning literature--the learning dynamics of infinitely wide neural networks under gradient descent with small step size is close to that of kernel models \cite{JacotHG18}.
Although this equivalence and error rates for the NTK approximation are known for general loss functions, such results are restricted to a few steps of gradient descent and not until convergence of training \cite{YangL21a}.
Only for specific losses (squared, logistic, the NTK approximation has been studied at convergence \cite{0001ZB20}.

Our interest in the NTK regime stems from the need to characterize the learned representation (discussed later). Hence, we are primarily concerned with the NTK approximation at convergence when minimizing self-supervised loss functions such as \eqref{eq:BT} or \eqref{eq:SCL}.
Below, we present the study in \cite{fleissner2025infinite}, which studies the case of Barlow twins loss function \eqref{eq:BT} with $\bC$ being the symmetrized cross-moment matrix \eqref{eq:symC} and $\lambda=1$. 
Observe that the loss function simplifies to $\mathcal{L}_{BT}(f) = \left \| \bC - \bI  \right \|^2$.
We restrict our analysis to the case where $f:\bbR^d\to\bbR^k$ is modeled by a two-layer neural network with hidden-layer width $M$,
\begin{align*}
    f(\bx) = \frac{1}{\sqrt{M}} \sum_{m=1}^M \bw_m \psi(\bv_m^\top \bx)\;,
\end{align*}
where $\bw_m \in \bbR^k$ and $\bv_m \in \bbR^d$ for all $m \in [M]$ are trainable parameters, and $\psi$ is a smooth bounded activation function with bounded first derivative. For example, we could have $\tanh$ activation. The weights are initialized as random independent Gaussians with constant variance, and collected in a vector $\btheta \in \bbR^{M(d+k)}$ that is trained under \emph{gradient flow} (gradient descent with infinitesimally small step sizes)
\begin{align*}
    \frac{\partial \btheta}{\partial t} = \dot{\btheta}(t) = - \frac{\partial \mathcal{L}_{BT}}{\partial \btheta}
\end{align*}
We write $\theta_0$ for the weights at initialization.
The neural tangent kernel (NTK) is defined as a time-varying, matrix-valued map \cite{JacotHG18}
\begin{align*}
    \bK_t (x,x') = \left( \left( \frac{\partial f_i(\bx)}{\partial \theta(t)} \right)^\top \left( \frac{\partial f_j(\bx')}{\partial \theta(t)} \right) \right)_{i,j=1}^k
\end{align*}
for all $\bx,\bx' \in \bbR^d$, where $f_i$ is the $i$-th output dimension of $f$. To underline the dependence of the NTK on the parameters $\theta(t)$ that evolve during training, we sometimes also denote it as $\bK_{\theta}$. The key insight of the NTK literature, proved mainly for squared loss \cite{JacotHG18,0001ZB20}, is that the NTK does not change during training if the width of the neural network approaches infinity. Consequently, the training dynamics of $f$ approach those of kernel regression with respect to the (vector-valued) kernel at initialization $\bK_0$. The constancy of the NTK in the infinite width limit essentially relies on three facts: 
\begin{enumerate}
    \item The spectral norm of the Hessian of the neural network is $\mathcal{O}\left(\frac{R}{\sqrt{M}}\right)$ for all weights $\theta$ with $\| \theta - \theta_0 \| \le R$.
    \item The change in the NTK from $\theta_0$ to any $\theta$ can be bounded in terms of the Hessian, and $\| \theta - \theta_0 \|$.
    \item $R$ is independent of $M$ because convergence happens in a ball of width-independent radius around $\theta_0$. 
\end{enumerate}
The first fact is true regardless of the loss function. The same is true for the second fact.
However, the third piece in the puzzle is missing: Unless $R$ remains independent of $M$, we do not obtain constancy of the NTK at a large width $M \rightarrow \infty$. To ensure that this holds, it is instructive to look at the evolution of the loss $\mathcal{L}_{BT}$ and the parameters \textit{ over time}. Defining
\begin{align*}
\begin{split}
    \bu(t) &= 
    \begin{bmatrix}
        \left( \frac{\partial \mathcal{L}_{BT}}{\partial f_j(\bx_i)} \right)_{i,j} \\
        \left( \frac{\partial \mathcal{L}_{BT}}{\partial f_j(\bx_i^+)} \right)_{i,j}
    \end{bmatrix}, \quad 
    \bK(t) = 
    \begin{bmatrix}
     \bK_t(x_1, x_1) & \dots & \bK_t(x_1, x_N^+) \\
     \vdots & \ddots & \vdots \\
     \bK_t(x_N^+, x_1) & \dots & \bK_t(x_N^+, x_N^+) \\
    \end{bmatrix}
\end{split}
\end{align*}
the time evolution of the Barlow Twins loss can be expressed as
\begin{align*}
    \frac{\partial}{\partial t} \mathcal{L}_{BT}(t) = - \bu(t)^\top \bK(t) \bu(t).
\end{align*}
Under certain assumptions, namely that the loss at initialization is smaller than $1$ and that the smallest eigenvalue of the kernel matrix at initialization is safely bounded away from zero, it can be shown that $\mathcal{L}_{BT}(t)$ decreases exponentially fast for \textbf{all} networks of sufficiently large width. Consequently, we may pick a time $T$ that is width-independent and ensures convergence of training until $T$. Additionally, it can be shown that for any $\epsilon > 0$, there exists a width-independent $\kappa>0$ such that $\sup_{t \le T} \| \dot{\theta}(t) \| \le \kappa$ holds with high probability $\ge 1 - \epsilon$ for any network of sufficiently large width, implying that the weights cannot have moved too much until convergence. Together, both of these observations yield the following result.

\begin{figure}[ht]
\centering
\includegraphics[width = 0.98\textwidth]{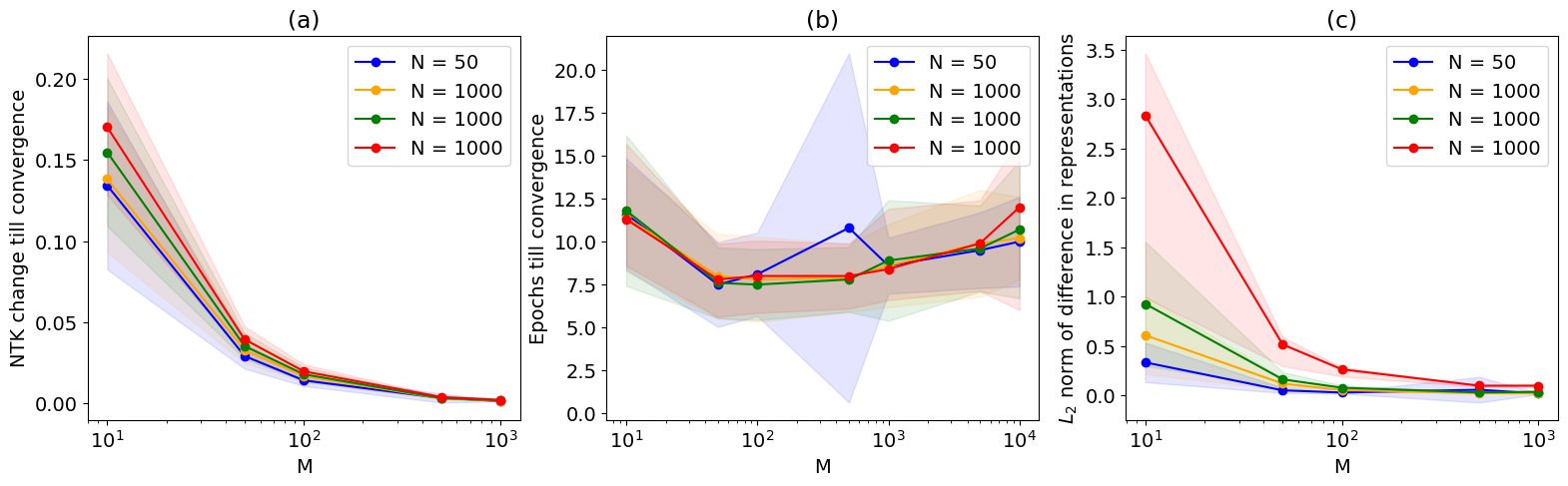}
\caption{Source \cite{fleissner2025infinite}. \textbf{Numerical evidence of constancy of NTK under Barlow Twins loss minimization.} 
We verify Theorem \ref{thm:NTK} by training a 1-hidden layer network with tanh activation on the MNIST dataset. We use gradient descent with a learning rate of 0.5, and train till loss $\mathcal{L}_{BT} \leq 10^{-5}$. The results are averaged over 10 independent runs.
For a fixed sample size $N$, we plot different quantities for varying network width $M$. We then vary $N$ and plot: (a) NTK change till convergence, where we see that as width increases, there is less change in NTK between initialization and convergence; (b) training epochs till convergence, which shows that the time to convergence remains almost constant with the network width; (c) squared norm of difference between representations of neural network and corresponding kernel model under the Barlow Twins loss, which validates that one can use the optimal solution of the kernel model (see Section 3.2) as a good approximation for the representation learned by a neural network.}
\label{fig:ntk_change}
\end{figure}

\begin{theorem}[Constancy of the NTK under Barlow Twins loss minimization \cite{fleissner2025infinite}]
\label{thm:NTK}
    Assume that for any network of sufficiently large width, the NTK matrix $\bK(0)$ is positive definite with the smallest eigenvalue $\lambda_{min}(\bK(0)) \ge \lambda > 0$, and that $\mathcal{L}_{BT}(0) \le 1-\rho < 1$ for some small $\rho>0$. Then, there exists a real number $R>0$ such that, with probability at least $1-\epsilon$, the change of the NTK until convergence of the loss (up to small $\delta>0$) is $\mathcal{O}(R^2/\sqrt{M})$. In particular, the radius $R$ depends only on $\delta, N, \lambda, \rho, \epsilon$, but not on the network width.
\end{theorem}
Figure \ref{fig:ntk_change} illustrates our theoretical results empirically: As the width of the network increases, the NTK changes less and less, and consequently the representations learned in a neural network approach those of a kernel model.

In this section, we consider only the training dynamics of networks under Barlow twins loss. It is natural to ask if a similar conclusion can be drawn for other self-supervised loss functions, for example VICReg or spectral contrastive loss.
In \cite{fleissner2025infinite}, we discuss how the analysis could be extended to other loss functions. A similar analysis for contrastive loss functions is presented in \cite{AnilEG25}, where it is also shown that the NTK equivalence does not hold for all loss functions. For a simple contrastive loss function, the output of the neural network diverges to infinity within a logarithmic number of gradient descent steps.
Our results on the convergence or divergence of training dynamics in \cite{fleissner2025infinite, AnilEG25} are particularly interesting because prior work implicitly assumes that the NTK equivalence always holds for self-supervised models. In particular, \cite{SimonKLGFA23} use the kernel equivalence to show that representations are learned in a stepwise manner, one dimension at a time, while \cite{CabannesKBLB23} used the kernel connection to study generalization properties and inductive biases of VICReg.

There also exist a few works beyond the NTK regime, focusing on specific settings.
\cite{WenL21,WenL22} study the learning dynamics of the joint embedding model with a 1-hidden layer ReLU network under sparse coding and strong/weak feature models and characterize the features learned. \cite{Tian22} study the dynamics of contrastive learning under coordinate-wise optimization. 

\subsection{Kernel-based self-supervised models learn spectral projections}

In this section, we focus on characterizing the learned representations $\bx \mapsto f(\bx)$, assuming that $f$ is a kernel model. 
Note that the NTK equivalence of trained neural networks at convergence, discussed in the previous section, allows us to interpret the subsequent discussion in the context of deep learning---we do not explicitly consider neural networks in this section.
Several works have characterized the optimal solution for various self-supervised loss functions, often assuming linear or kernel models. The essence of such a characterization could be summarized as \emph{``optimal representations learned by self-supervised models are spectral projections''}. However, the results differ somewhat in terms of the matrix or operator whose spectral decomposition is used.
The most notable work is \cite{BalestrieroL22}, which provides a unified framework that relates several self-supervised loss functions and argues that any of the loss minimization is related to spectral embedding.
\cite{SimonKLGFA23,CabannesKBLB23} provide more precise characterization of Barlow Twins and VICReg loss minimization, respectively, in terms of spectral decomposition of a cross-covariance operator.
In \cite{EsserFG24}, we provide a corresponding result for several contrastive loss functions, including the spectral contrastive loss.
Using a slightly different perspective, \cite{HaoChen2021ProvableGF,0001HM23} show that joint embedding methods learn spectral projections of a \emph{augmentation graph}---a graph over all unlabeled augmented data with edges connecting the augmented pairs.

In line with the earlier discussions on the Barlow Twins loss, we state a characterization of the representations learned by minimizing $\mathcal{L}_{BT}$, adapted from \cite{SimonKLGFA23}. 
Recall that the loss is given by $\mathcal{L}_{BT}(f) = \| \bC - \bI_k \|^2$, where $\bC \in \bbR^{k \times k}$ is the empirical cross-moment matrix of $f(\bx),f(\bx^+)$. 
But under a kernel model $f(\bx) = \bW \phi(\bx)$, it can be rephrased as
\begin{align*}
    &\mathcal{L}_{BT}(\bW) = \| \bW \Gamma \bW^\top - \bI_k \|^2, \quad \text{where~ } 
    \Gamma = \frac{1}{2N} \sum_{i=1}^{N} \phi(\bx_i) \phi(\bx_i^+)^\top + \phi(\bx_i^+) \phi(\bx_i)^\top 
\end{align*}
is the symmetrized cross-moment matrix of the feature map $\phi:\bbR^d \to\mathcal{H}$.
Note that we use the notation $\ba\bb^\top$ in the sense of an outer product. 
The following result holds.
\begin{theorem}[Optimal representations learned by kernel Barlow Twins model, adapted from \cite{SimonKLGFA23}]
    \label{thm: optimal representation}
    Let the eigen pairs of $\Gamma$ be given by $(\lambda_i, \bu_i)_{i=1,2,\ldots}$ in decreasing order of eigenvalues.
    If $\lambda_k > 0$, the projection $\bW^* : \mathcal{H} \to \bbR^k$ given by 
    $\displaystyle \bW^* = \bQ \left[ \begin{array}{c} 
        \frac{1}{\sqrt{\lambda_1}}\bu_1^\top
        \\ \vdots \\ 
        \frac{1}{\sqrt{\lambda_k}}\bu_k^\top
        \end{array}\right]$,
    where $\bQ \in \bbR^{k \times k}$ is any orthogonal matrix,
    achieves $\mathcal{L}_{BT} =0$. 
    In particular, $\bW^*$ is the minimum norm solution that achieves zero loss, and under some conditions on initialization, the gradient descent converges to the solution $\bW^*$.

    The learned representation can be expressed in terms of the kernel function $\kappa$ in the following way. 
    Let $\mathcal{X} = \{\bx_1,\ldots,\bx_N\}$ and $\mathcal{X}^+ = \{\bx_1^+,\ldots,\bx_N^+\}$ refer to the unlabeled data and their respective augmentation, and $\bx$ be the new sample for which we compute $f(\bx)= \bW^*\phi(\bx)$.
    We use $\bK_{\mathcal{X}\mathcal{X}^+} \in \bbR^{N\times N}$ denote a matrix with $\left[\bK_{\bx\mathcal{X}}\right]_i = \kappa(\bx_i, \bx)$ and analogously define $\bK_{\mathcal{X}\mathcal{X}}, \bK_{\mathcal{X}^+\mathcal{X}^+}, \bK_{\mathcal{X}^+\mathcal{X}} \in \bbR^{N\times N}$ and $\bK_{\mathcal{X}x},\bK_{\mathcal{X}^+x} \in \bbR^{N\times 1}$.
    Define the matrices
    \begin{align*}
    \bZ &= \frac{1}{2N}\left(
        \left[ \begin{array}{c} 
        \bK_{\mathcal{X}\mathcal{X}^+}
        \\
        \bK_{\mathcal{X}^+\mathcal{X}^+}
        \end{array}\right]
        \left[ \begin{array}{cc} 
        \bK_{\mathcal{X}\mathcal{X}}
        &
        ~\bK_{\mathcal{X}\mathcal{X}^+}
        \end{array}\right]
        +
        \left[ \begin{array}{c} 
        \bK_{\mathcal{X}\mathcal{X}}
        \\
        \bK_{\mathcal{X}^+\mathcal{X}}
        \end{array}\right]
        \left[ \begin{array}{cc} 
        \bK_{\mathcal{X}^+\mathcal{X}}
        &
        ~\bK_{\mathcal{X}^+\mathcal{X}^+}
        \end{array}\right]
        \right),
    \\
    \bK &= \left[ \begin{array}{cc} 
        \bK_{\mathcal{X}\mathcal{X}} & ~\bK_{\mathcal{X}\mathcal{X}^+}  
        \\
        \bK_{\mathcal{X}^+\mathcal{X}} & ~\bK_{\mathcal{X}^+\mathcal{X}^+}  
        \end{array}\right]
    \qquad\text{and}\qquad
    \bK_\Gamma = \bK^{-1/2}\bZ\bK^{-1/2}.
    \end{align*}
    The learned representations are given by
    \begin{align*}
    f(\bx) = \bW^*\phi(\bx) = \bQ 
    \left[ \begin{array}{c} 
        \frac{1}{\sqrt{\lambda_1}}\bv_1^\top
        \\ \vdots \\ 
        \frac{1}{\sqrt{\lambda_k}}\bv_k^\top
        \end{array}\right]
    \bK^{-1/2}
    \left[ \begin{array}{c} 
        \bK_{\mathcal{X}\bx}
        \\
        \bK_{\mathcal{X}^+\bx}
        \end{array}\right],
    \end{align*}
    with $\bQ \in \bbR^{k\times k}$ any orthogonal matrix and 
    $(\lambda_i,\bv_i)$ eigen pairs of $\bK_\Gamma$ (which has same nonzero eigenvalues as $\Gamma$).
\end{theorem}

The first part of the theorem asserts that diagonalizing the cross-moment matrix $\Gamma$ is done best (in a least norm sense) by projecting onto the top eigenvectors, which is also the solution learned by gradient descent due to \emph{implicit bias}. Note that if $\lambda_k\le 0$, it is impossible to achieve zero loss, since $\bW \Gamma \bW^\top$ cannot have more positive eigenvalues than $\Gamma$ itself.
The second part of the result provides the practically implementable representation in terms of the kernel function $\kappa$ using the so-called \emph{kernel trick}. 
Note that the resulting solution is bit more complicated in comparison to kernel ridge regression or kernel PCA. This is because the Barlow Twins loss $\mathcal{L}_{BT}$ is \emph{quartic} (degree 4) in the parameters $\bW$ instead of quadratic objectives of PCA or squared regression.

To summarize the above discussion, we see that in a kernel setting, it becomes possible to precisely characterize the (minimum norm) solutions of Barlow Twins, which interestingly reveals that $\mathcal{L}_{BT}$ essentially projects onto the dominant eigen space of the cross-moment matrix between positive pairs $(\bx,\bx^+)$, thereby generalizing classical unsupervised representation learning methods like kernel PCA.
This underlines the vital role of the augmentation in determining the patterns learned in joint embedding methods, which we investigate next.

\subsection{Optimal data augmentation to express desired representation}

The choice of augmentations is crucial to the performance of self-supervised representation learning, since different downstream tasks may require dramatically different augmentations. The literature on visual foundation models contains several empirical works that study which augmentations work better and how many different augmentations are needed (see the discussions in \cite{feigin2024theoretical}). For example, it is known that while cropping encourages invariance to occlusions, it also negatively affects downstream tasks that require category and viewpoint invariance \cite{purushwalkam2020demystifying}. There is limited theoretical understanding regarding the subtleties of the augmentation choice. 

In particular, a fundamental question remains unanswered:
\emph{What is the relationship between data, loss function, and the augmentations needed to learn a desired target representation $f^*$?} 
Observe that the question is similar to the notion of \emph{expressivity} or \emph{universal approximation} \cite{DeVoreHP21}. However, unlike supervised learning, where the model has access to the target representation through labeled data $\big\{(\bx_i,f^*(\bx_i))\big\}_{i=1}^N$, the above question asks if one can learn $f^*$ only with access to unlabeled augmented pairs $\left\{(\bx_i,\bx_i^+)\right\}_{i=1}^N$.

We address the above question in \cite{feigin2024theoretical},
which also makes practical considerations by allowing random data augmentation. For ease of exposition, we consider a simpler setup of deterministic augmentation and adapt the results of \cite{feigin2024theoretical} accordingly.
\begin{quote}

\textbf{Formal problem on expressivity of self-supervised models.}
Given unlabeled data $\bx_1,\ldots,\bx_N \in \bbR^d$ and a target representation $f^*:\bbR^d\to\bbR^k$, does there exist an augmentation map $\bx_ \mapsto \bx^+ = T(\bx)$ such that the  representation $f$ learned by minimizing $\mathcal{L}_{BT}$ \eqref{eq:BT} or $\mathcal{L}_{SCL}$ \eqref{eq:SCL} using the augmented pairs $\big\{(\bx_i,T(\bx_i))\big\}_{i=1}^N$ is identical to $f^*$, up to affine transformation?
\end{quote}
Note that it suffices to learn $f^*$ up to the affine transformation since, in practice, a linear predictor is typically fitted using the learned embedding $f^*(\bx)$ for downstream prediction tasks (see, for example, Figure \ref{fig:Emergent}).
Assuming that $f$ is learned using a kernel self-supervised model, we can exploit the characterization of the optimal representation (such as Theorem \ref{thm: optimal representation}) to derive the optimal data augmentation.
To this end, we relax the above-mentioned problem and try to prove the existence of a transformation $T_\mathcal{H}:\mathcal{H}\to\mathcal{H}$, $\phi(\bx_i) \mapsto \phi_i^+ \in \mathcal{H}$ instead of finding the augmentation in the input space $\bx_i \mapsto \bx_i^+$, where $\mathcal{H}$ in the reproducing kernel Hilbert space (rkhs) $\mathcal{H}$ associated with the kernel $\kappa$. \cite{feigin2024theoretical} characterizes $T_\mathcal{H}$ for $\mathcal{L}_{BT}$ and $\mathcal{L}_{SCL}$, which we state below.

\begin{theorem}[Optimal data augmentation for spectral contrastive and Barlow Twins models \cite{feigin2024theoretical}]
\label{thm: optimal augmentation}
Let $\bx_1,\ldots,\bx_N \in\bbR^d$ be given unlabeled data, $f^*:\bbR^d \to \bbR^k$ be a target representation, and $\kappa: \bbR^d \times \bbR^d \to \bbR$ be a pre-specified kernel with associated rkhs $\mathcal{H}$ and map $\phi:\bbR^d\to\mathcal{H}$.
Define $\Phi = [\phi(\bx_1) ~\cdots~ \phi(\bx_N)]$.

Assume that $f^* = \bS \Phi^\top$ and that both the kernel matrix $\bK = \Phi^\top \Phi \in \bbR^{N\times N}$ and the matrix $\bS\bK\bS^\top \in \bbR^{k \times k}$ have full rank.
The following statements hold.
\begin{itemize}
    \item \textbf{Spectral constrastive:} 
    Define the augmentation $T_\mathcal{H}:\mathcal{H}\to\mathcal{H}$ such that for $\bx \in \bbR^d$, 
    \[ 
    \phi^+ = T_\mathcal{H}(\phi(\bx)) = \Phi\,\bS^\top (\bS\,\bK\,\bS^\top)^{-1}\bS\,\Phi^\top \phi(\bx).
    \]
    If $f$ is the least norm minimizer of $\mathcal{L}_{SCL}$ \eqref{eq:SCL} with additional regularization $\frac1N \sum\limits_{i=1}^N \Vert f(\bx_i)\Vert^2 + \Vert f(\bx_i^+)\Vert^2$ and augmented pairs $\left\{\left(\phi(\bx_i),\phi_i^+\right)\right\}_{i=1}^N$, then $f = f^*$ up to affine transformation. 
    \item \textbf{Barlow Twins:} Define the augmentation $T_\mathcal{H}:\mathcal{H}\to\mathcal{H}$ such that for any $\bx \in \bbR^d$, 
    \[ 
    \phi^+ = T_\mathcal{H}(\phi(\bx)) = \Phi \bK^{-1/2}\bB \bK^{-1/2}\Phi^\top \phi(\bx),
    \]
    where  $\bB$ is a solution to the Lyapunov equation
    \[
    \bK\bB + \bB\bK^\top =2N \cdot \bK^{1/2} \bS^\top \left(\bS\bK\bS^\top \right)^{-2}\bS\bK^{1/2}
    \]
    If $f$ is the least norm minimizer of $\mathcal{L}_{BT}$ \eqref{eq:BT} with augmented pairs $\left\{\left(\phi(\bx_i),\phi_i^+\right)\right\}_{i=1}^N$, then $f = f^*$ up to affine transformation. 
\end{itemize}
\end{theorem}

It is important to discuss the assumptions made in the above theorem. The assumption that $\bK$ is full rank holds for any universal kernel (for example, Gaussian, Laplace, etc.) and distinct points $\bx_1,\ldots,\bx_N$.
Assuming that the target $f^*=\bS\Phi^\top $, or rather that $f^*\in \text{span}\big\{\phi(\bx_1),\ldots,\phi(\bx_N)\big\}$ ensures that $f^*$ is itself a minimum norm solution since, by virtue of  the representer theorem \cite{ScholkopfMIT}, any least norm minimizer of a loss function is contained in the span.
One could extend the theorem to the cases $f^*\notin \text{span}\big\{\phi(\bx_1),\ldots,\phi(\bx_N)\big\}$ or $f^* \notin \mathcal{H}$ by accounting for the error in projecting on to the span, which can be made arbitrarily small by using universal kernels and allowing arbitrarily large $N$. 
Finally, the full rank of $\bS\bK\bS^\top$ implies that the sample covariance of $\{f^*(\bx_i)\}_{i=1}^N)$ is full rank. 

\begin{figure}[ht]
\centering
\fbox{\includegraphics[height = 0.28\textwidth,clip=tru,trim=5mm 0mm 10mm 4mm]{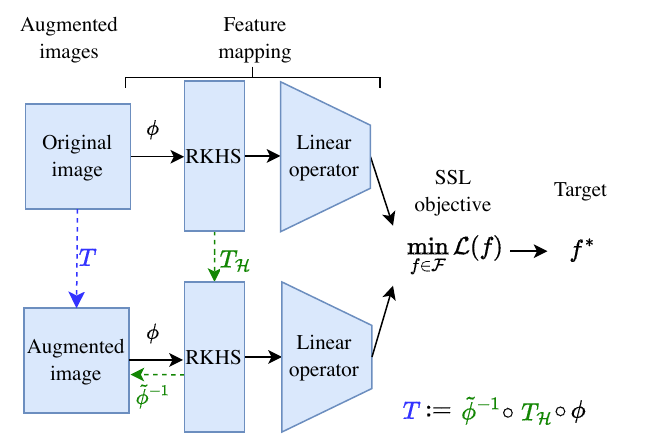}}
~~\includegraphics[height = 0.28\textwidth]{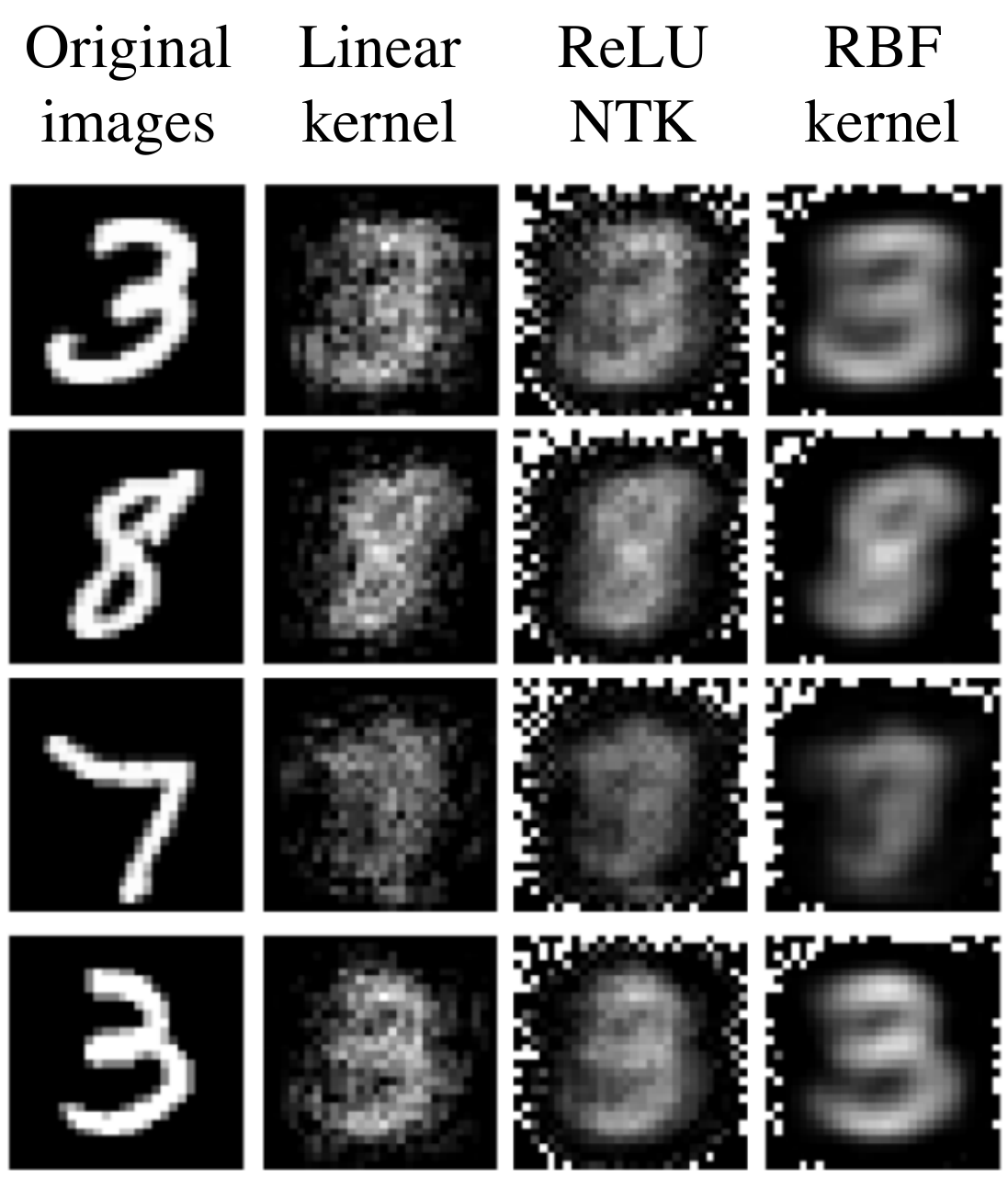}
~~\includegraphics[height = 0.28\textwidth]{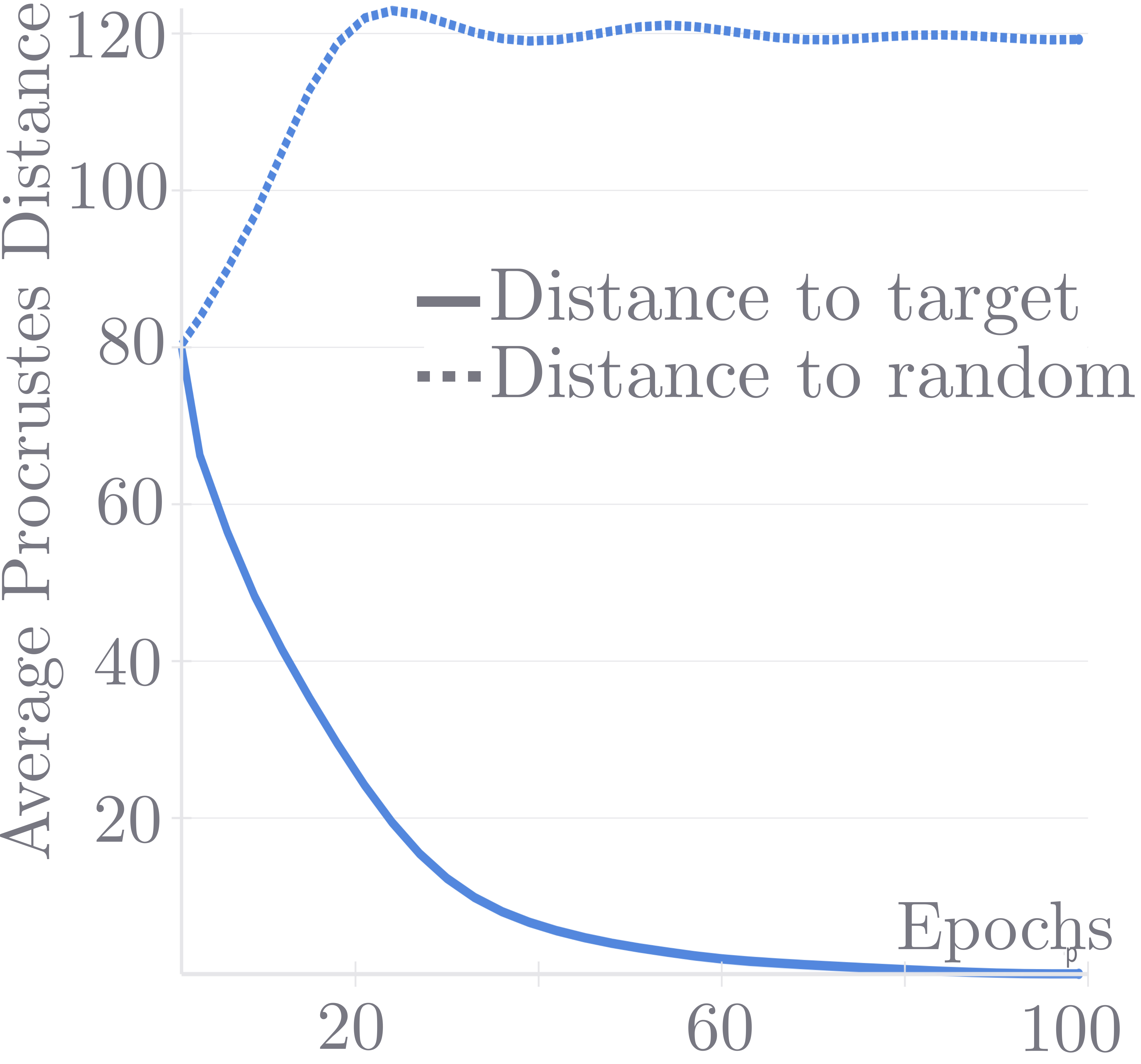}
\caption{Source \cite{feigin2024theoretical}. \textbf{Optimal data augmentation for kernel spectral contrastive model.} 
The left plot shows a schematic for constructing augmentation $\bx \mapsto \bx^+$ by using the optimal augmentation $T_\mathcal{H}$ in the rkhs $\mathcal{H}$.
The middle plot compares augmented MNIST images using different kernels, demonstrating different function classes require different augmentations to achieve the same representations. Here, the target representation $f^*$ is considered as the one obtained from a pre-trained ResNet50. 
The right plot shows that average Procrustes distance between the learned representation $f$ from target $f^*$ and random representations during training given the augmentations $\mathcal{T}_\mathcal{H}$. It validates our theory that the analytically derives $T_\mathcal{H}$ indeed results in learning the target representation $f^*$.
     }
\label{fig:optimal augment}
\end{figure}

To summarize the above discussion, we show that under certain conditions, a target $f^*$ can be learned through self-supervised learning using specific augmentations in the rkhs. It is easy to see that, in the context of spectral contrastive models, the augmentation $T_\mathcal{H}$ is simply a rank-$k$ projection in $\mathcal{H}$.
It is natural to ask if one can derive the corresponding augmentation $\bx \mapsto \bx^+$ in the input space.
This may not be possible, in general, as the map $\phi:\bbR^d\to\mathcal{H}$ may not be surjective. 
However, \cite{feigin2024theoretical} presents an approach to numerically find approximate augmentations in the input space by solving a \emph{kernel pre-image problem}. Figure \ref{fig:optimal augment} illustrates the optimal augmentation for different kernel models, which are surprisingly quite different from commonly used augmentations like rotation or cropping. 

\subsection{Statistical generalization for learned representations}

We return to the topic of statistical generalization, which is the central theme of this paper. The limiting kernel regime for self-supervised neural networks as well as the characterization of the learned representations in the kernel setting help to derive generalization error bounds for joint embedding models.
Note that the question of generalization needs to be studied in two separate contexts. In this section, we study the generalization of the learned representation to new samples with respect to a self-supervised loss, \eqref{eq:BT} or \eqref{eq:SCL}. In the next section, we discuss the pertinent topic of the generalization error of downstream predictors.

In supervised settings or the case of reconstruction, it makes sense to bound the generalization error defined as the expected loss on a new sample. However, this notion of generalization may not be meaningful in the context of self-supervised losses. For example, consider the case of simplified Barlow Twins loss $\mathcal{L}_{BT}(f) = \Vert \bC - \bI_k\Vert^2$ and assume that there is an underlying distribution for the augmented pairs $(\bx,\bx^+)$. The natural notion of generalization error in this case would be $\bbE_{(\bx,\bx^+)}\left[ \left\Vert \frac12\left(f(\bx)f(\bx^+)^\top + f(\bx^+)f(\bx)^\top\right) - \bI_k\right\Vert^2\right]$, which is always at least $k-2$ since the cross-covariance term has rank 2.
In \cite{fleissner2025infinite}, we avoid this issue by deriving error bounds for a different notion of a population loss for Barlow Twins.

It is possible to state generalization error bounds for the spectral contrastive loss \eqref{eq:SCL} using classical uniform convergence results, as we state below.
\begin{theorem}[Generalization error bound for kernel spectral contrastive model, adapted from \cite{EsserFG24}]
\label{th:generalisation SCL}
Assume that the kernel $\kappa$ is bounded with $\nu := \sup_{\bx} \kappa(\bx,\bx)$ and the augmented pairs $\left\{(\bx_i,\bx_i^+)\right\}_{i=1}^N$ are independent and identically distributed samples from a distribution $\mathcal{D}$ on $\bbR^d\times\bbR^d$.

Let $\widehat{f} = \underset{f\in\mathcal{F}}{\text{arg\,min}} ~\mathcal{L}_{SCL}(f)$, where $\mathcal{L}_{SCL}(f)$ is the spectral contrastive loss in \eqref{eq:SCL} and the optimization is over the hypothesis class of kernel models with a bounded norm $\mathcal{F} = \big\{ f(\bx) = \bW \phi(\bx) \;:\; \Vert\bW \Vert \leq \omega\big\}$.
Let $\mathcal{L}'_{SCL}$ denote the loss for an independent set of pairs $\left\{(\bz_i,\bz_i^+)\right\}_{i=1}^n \sim \mathcal{D}^n$. With probability $1-\delta$
\[
\bbE_{\{(\bz_i,\bz_i^+)\}_i\sim \mathcal{D}^n} \left[\mathcal{L}'_{SCL}(\widehat{f})\right] \leq \mathcal{L}_{SCL}(\widehat{f}) + O\left(  \omega^3\kappa^2\sqrt{\frac{k + \omega^2\log\frac{1}{\delta}}{N}}\right).
\]
\end{theorem}

\section{Statistical performance guarantees for downstream predictors}

In the modern context of foundation models, unsupervised representation learning as a \emph{pretraining} phase.
Good representations help reduce the number of labeled samples required to train supervised predictors.
A well-known example of this is \emph{principal component regression}, where one first uses PCA to reduce the dimension of training data and then fits the regressor. PCA has been established to act as a regularizer for regression, specifically when the true signal is aligned with the principal components \cite{DickerFH17}. 
Hence, statistically, alternative (self-supervised) approaches are needed when the ideal representations for downstream prediction do not align with the principal components of the unlabeled data. 
Theorem \ref{thm: optimal augmentation} establishes that, with appropriate augmentation, we can learn any target representation. Hence, the general strategy for deriving statistical guarantees for downstream generalization is to integrate the aforementioned results into a statistical framework.

In recent years, two distinct principles have been used to derive downstream generalization error bounds. 
\cite{Arora2019ATA} introduced a \emph{contrastive unsupervised representation learning (CURL)} framework, which can be applied to general joint embedding methods. Here, one assumes that the augmented unlabeled data is sampled from a mixture distribution with augmented pairs having a higher probability of being sampled from the same class. 
\cite{HaoChen2021ProvableGF} propose an alternative framework that presents generalization error bounds in terms of properties of a population augmentation graph, where edges connect samples from same class. 
Despite being more general than CURL, the augmentation graph-based framework is limited, since the population and the empirical augmentation graphs are not known to be close. We are also unaware of tight error bounds that arise from this framework. 
Hence, in this section, we present results based on the CURL framework \cite{Arora2019ATA,vanElstG25}.

The statistical framework in CURL \cite{Arora2019ATA} is formalized as follows. Assume that there are $p$ classes with respective class conditional distributions $\mathcal{D}(\cdot|y)$ on $\bbR^d$, where $y \in [p]$. It is assumed that each augmented pair $(\bx,\bx^+)$ consists of samples of the same class. 
Using the above setup, one can derive uniform convergence bounds \cite{Arora2019ATA,EsserFG24} or tighter PAC-Bayesian bounds \cite{vanElstG25} on the self-supervised loss---the latter bounds also avoid the independence assumption about samples in augmented pair.
Once a representation $f:\bbR^d\to\bbR^k$ is learned by pre-training, a linear classifier is assumed to be trained for  label prediction.
The idea is to relate the downstream prediction error to the self-supervised loss, where the tightness of the bounds depends on the losses considered.
We present a result from \cite{vanElstG25} that relates the SimCLR contrastive loss \cite{ChenK0H20} to the supervised cross-entropy loss. We first define the two loss functions.
For augmented data $\left\{(\bx_i,\bx_i^+)\right\}_{i=1}^N$, the SimCLR loss is given by
\begin{equation}
\mathcal{L}_{SimCLR}(f) = \frac1N \sum_{i=1}^N -\log \frac{\exp \left(\frac{f(\bx_i)^\top f(\bx_i^+)}{\tau}\right)}{\exp \left(\frac{f(\bx_i)^\top f(\bx_i^+)}{\tau}\right) + \sum_{j \neq i} \exp \left(\frac{f(\bx_i)^\top f(\bx_j^+)}{\tau}\right)},
\label{eq:SimCLR}
\end{equation}
where $\tau>0$ is a temperature scaling hyperparameter that significantly impacts performance (see Table \ref{tab:risk_certificates_combined}).
On the other hand, for a labeled sample $(\bx,y)\in \bbR^d\times[p]$ and a learned embedding $f:\bbR^d\to\bbR^k$, assume that a $p$ class classifier is characterized by the matrix $\bA = [a_1, \ldots a_p] \in \bbR^{k \times p}$, where the prediction for class-$j$ uses the projection $f(\bx)^\top a_j$. The cross entropy loss is given by
\begin{equation}
    \label{eq:CE}
     \mathcal{L}_{CE}(f, \bA) =-\log \frac{\exp \left(f(\bx)^\top a_y\right)}{\sum_{y'=1}^p \exp \left(f(\bx)^\top a_{y'}\right)}\;.
\end{equation}

\begin{theorem}[Generalization error bounds for supervised cross entropy loss in terms of SimCLR loss \cite{vanElstG25}]
\label{thm:downstream}
Suppose the labeled data distribution $\mathcal{D}$ on $\bbR^d\times[p]$ is characterized by $\pi_j := \bbP_{(\bx,y)\sim\mathcal{D}}(y=j)$.  
Define $\Delta = \log\left(\frac{p}{N-1}\operatorname{cosh}^2\left(\frac1\tau\right)\right)$ and $\alpha = \log p + \min\{0, \log(\operatorname{cosh}^2(1)) - \tau \Delta\}$.
Then for any representation $f:\bbR^d\to\bbS^{k-1}$ (unit sphere in $\bbR^k$),
\begin{align*}
    \min _{\bA \in \mathbb{R}^{k \times p}} &\bbE_{(\bx,y)\sim\mathcal{D}} \left[\mathcal{L}_{CE} (f, \bA)\right] 
    \\&\leq  \min 
    \left\{\frac{\sigma}{\tau} + \Delta + \bbE\left[\mathcal{L}_{SimCLR}(f)\right], \sigma + \tau\Delta + \alpha + \tau \bbE\left[\mathcal{L}_{SimCLR}(f)\right] \right\}.
\end{align*}
Here, $\sigma = \bbE_{(\bx,y)\sim\mathcal{D}} \left\Vert f(\bx) - a^*_y\right\Vert \leq 2$, where $a^*_y$ denotes the population mean of the representation $f$, conditioned on label $y$. The expectation of $\mathcal{L}_{SimCLR}$ is with respect to the unlabeled samples used in pre-training.
\end{theorem}

The above result shows that the cross-entropy generalization error of the optimal linear classifier is bounded by the population SimCLR loss along with few additional terms.
It is natural to question the utility of Theorem \ref{thm:downstream}. In practice, the bound can be incorporated into a PAC-Bayesian framework to derive risk certificates (see the numerical results in Table \ref{tab:risk_certificates_combined}).
As shown in \cite{bao2022surrogate}, such bounds lead to non-vacuous error bounds that reflect the general trends of the test error much better than uniform convergence bounds (see Figure 1 in \cite{bao2022surrogate}).
Theorem \ref{thm:downstream} is a refinement of \cite{bao2022surrogate} and is tighter for a low temperature $\tau$.

\begin{table}[ht]
\centering
\scriptsize 
\setlength{\tabcolsep}{3pt} 
\resizebox{\textwidth}{!}{%
\begin{tabular}{c@{\hspace{1mm}}c@{\hspace{1mm}}ccc|c@{\hspace{1mm}}cccc}
\toprule
& & \multicolumn{3}{c}{SimCLR loss} 
    && \multicolumn{4}{c}{Cross entropy (supervised)}\\
 & & $\tau=1.0$ & $\tau=0.5$ & $\tau=0.2$ 
    && Test loss & $\tau=1.0$ & $\tau=0.5$  & $\tau=0.2$ \\
\midrule
 & Test loss & 4.9 & 4.26 & 2.71
    && 1.76 & 1.70 & 1.70 \\
\midrule
\multirow{3}{*}{\rotatebox{90}{Bounds}} 
 & PAC-Bayes i.i.d. & 8.48 & 8.91 & 10.17 
    && \cite{bao2022surrogate} & \textbf{3.27} & 5.30  & 12.59 \\
 & $f$-divergence \cite{nozawa2020pac} & 27.03 & 33.27 & 48.47
    && \cite[Theorem 5]{vanElstG25} & \textbf{3.27} &  \textbf{4.95} & \textbf{4.61} \\    
 & \cite[Theorem 3]{vanElstG25} & \textbf{5.20} & 6.27 & 43.77 
    && \\
 & \cite[Theorem 4]{vanElstG25} & 5.54 & \textbf{5.49} & \textbf{6.22}
    && \\
\bottomrule
\end{tabular}%
}
\caption{Source \cite{vanElstG25}. \textbf{Tight PAC-Bayesian generalization error bounds for the SimCLR loss and downstream supervised cross-entropy loss.} The empirical results and the bounds are for a 7-layer convolutional network pretrained on CIFAR-10 data set. 
The columns on the left show the empirical SimCLR loss and respective theoretical bounds using classical PAC-Bayesian bounds with the incorrect assumption that all samples are independent, $f$-divergence based PAC-Bayes bounds from \cite{nozawa2020pac} and our bounds \cite{vanElstG25} using a McAllester-type and KL-divergence-based PAC-Bayes bounds. The latter three bounds account for sample dependence. The temperature scaling $\tau$ heavily influences the empirical SimCLR loss and only one of bounds are non-vacuous at small temperature.
The columns on the right show the downstream test error and generalization error bounds for the cross-entropy loss. Our bounds \cite{vanElstG25} improve up existing bounds at small temperature.}
\label{tab:risk_certificates_combined}
\end{table}

\section{Future directions and open questions}

The statistical theory of unsupervised representation learning is still in a nascent stage. Compared with supervised deep learning theory, this direction of research is more complex and interesting because of the inherent challenges of formalizing unsupervised learning. However, there is significantly less attention to this topic from the machine learning theory community. The most surprising fact is that the most recent theoretical advances related to statistical generalization in foundation models, such as \cite{YangH21,BahriEKLS24}, mainly focus on supervised settings and do not address open questions about unsupervised representation learning or generalization guarantees for downstream prediction.
This paper attempts to provide an overview of recent theoretical research on unsupervised (deep) representation learning. Although many of the aforementioned results are taken from our recent work, there are also other interesting works on this topic, some of which are cited above.
We conclude this paper with some important open questions about this topic, beginning with the crucial question of precise error rates.
\begin{quote}

\textbf{Optimal generalization error rates.}
One of the main goals in statistics is to characterize minimum possible error rates and derive the optimal methods. Although Theorems \ref{th:generalisation SCL}--\ref{thm:downstream} provide upper bounds on self-supervised or downstream generalization errors, they do not provide a precise characterization like Theorems \ref{theorem:test_risk_noskip}--\ref{theorem:test_risk_skip}. In particular, lower bounds of the generalization error are needed in the context of foundation models. 
Recent work \cite{DengHZM24,GeTF024} has taken key steps in studying self-supervised learning from the perspective of meta learning and maximum likelihood estimation, which provide a statistical framework to study lower bounds of error and lead to the design of optimal approaches.
We also believe that the kernel self-supervised model described in this paper can help to derive precise error rates similar to the results on principal component regression \cite{DickerFH17}.  
\end{quote}
The other notable open direction is to understand the role of neural networks in unsupervised representation learning. We may formally pose the following questions.
\begin{quote}
    \textbf{Role of feature and attention learning in unsupervised models.}
    Recent literature on supervised learning has studied the statistical advantage of using deep neural networks or transformer architectures. For example, in the context of learning a $k$-parity function, \cite{DanielyM20,HanG25} proves that, compared to kernel models, trained 1-hidden-layer networks or transformers require significantly fewer parameters.
    In a similar vein, one could ask whether training a non-linear autoencoder or joint embedding model provably recovers the signal better than kernel models.
\end{quote}
A more precise understanding of unsupervised representation learning and its impact on downstream predictors will not only help in the theory of foundation models but also significantly advance the fields of statistics and theoretical machine learning as a whole.

\paragraph{\bf Acknowledgment and funding information.}
The research contribution of P. M. Esser, mentioned in this paper, was carried out when he was at the Technical University of Munich, supported by the German Research Foundation (DFG) Priority Program 2298 ``Theoretical Foundations of Deep Learning'' (project GH-257/2-1). 
M. Fleissner is supported by the DAAD program Konrad Zuse Schools of Excellence in Artificial Intelligence, sponsored by the Federal Ministry of Education and Research (BMBF).
In addition, D. Ghoshdastidar acknowledges the German Research Foundation for funding through the project GH-257/4-1, DFG-ANR PRCI ``ASCAI'' (project GH-257/3-1) and DFG GRK 2428 ``ConVeY''. 
The authors thank Gilles Blanchard, Leena Chennuru Vankadara, Satyaki Mukherjee, Mahalakshmi Sabanayagam, Maedeh Zarvandi, Gautham Govind Anil, Shlomo Libo Feigin, Jonghyun Ham, Anna van Elst and Theresa Wasserer, who have collaborated on some of the cited works or are currently collaborating on this topic.

\printbibliography 

\end{document}